# AEF-Econ: Toward Plug-and-Play Socioeconomic Foundation Embeddings from AlphaEarth for Urban Remote Sensing

Shuyang Hou[a]*, Ziqi Liu[a]*, Haoyue Jiao[a], Lutong Xie[a], Yaxian Qing[a], Xiaopu Zhang[a], Qingyang Xu[a], Zhangyan Xu[a], Xuefeng Guan[a], Huayi Wu[a]

*a. State Key Laboratory of Information Engineering in Surveying, Mapping, and Remote Sensing, Wuhan University, Wuhan, China

*Corresponding author: Ziqi Liu (first corresponding author), email: lzq677@whu.edu.cn

Shuyang Hou (second corresponding author), email: whuhsy@whu.edu.cn

## Abstract

AlphaEarth Foundations (AEF) unify global remote sensing foundation embeddings through multimodal self-supervised learning, but their pretraining focuses on physical land-surface signals, limiting plug-and-play use in socioeconomic tasks. We integrate seven heterogeneous data streams across 36 Chinese cities over eight years—AEF embeddings, population, nighttime lights, remote sensing indices, points of interest (POIs), urban morphology, and cross-lingual text—and construct CHN-Econ, a socioeconomic benchmark with 16 labels in three categories. We conduct 31 controlled experiments along five axes: fusion architecture, self-supervised objective, text integration, embedding dimensionality, and normalization. Used alone as a linear probe, AEF achieves $R^2$ values of only 0.301 for cross-region and 0.160 for cross-tier evaluation. The five-axis ablated backbone improves these scores to 0.832 and 0.671, respectively, but reveals that low-dimensional semantic streams are consistently suppressed by high-dimensional streams under shared reconstruction. To address this bottleneck, we propose Capacity-Adaptive Reconstruction (CAR), replacing shared reconstruction with per-stream decoders and stream-level losses to mitigate inter-stream capacity competition. CAR further raises cross-region and cross-tier $R^2$ to 0.848 and 0.693, and restores collapsed labels from negative $R^2$ to a stable range. Using CAR, we infer 14.4 million pixels across 36 cities and eight years and release AEF-Econ, including 128d and 64d compressed versions. Self-diagnostics and case studies show that AEF-Econ captures cross-city hierarchies and intra-urban spatial organization under unsupervised settings, providing a socioeconomic remote sensing foundation embedding complementary to AEF physical embeddings.

## 1. Introduction

Remote sensing Earth observation continuously records land-surface changes at high spatiotemporal resolution, providing an important technical pathway for socioeconomic research to overcome the temporal lag of conventional statistical yearbooks (Jean et al., 2016). However, traditional task-specific supervised paradigms face three major limitations (Hou et al., 2026d). First, models are typically designed for a single task and have limited cross-task reusability. Second, multi-source fusion often depends on heuristic feature concatenation rather than a unified representation space. Third, generalization across regions and development levels remains insufficient. In recent years, self-supervised representation learning has emerged as a promising paradigm, with representative studies including MOSAIKS (Rolf et al., 2021), SatMAE (Cong et al., 2022), Prithvi (Jakubik et al., 2023), SatCLIP (Klemmer et al., 2023), DOFA (Xiong et al., 2024), Tessera (Feng et al., 2025), and AlphaEarth Foundations (AEF) (Brown et al., 2025). Among them, AEF takes nine categories of global gridded physical sources and English Wikipedia text as inputs to generate annual 10-m global embeddings for 2017–2024. Across 15 cross-domain tasks, it reduces errors by approximately 23.9% on average relative to existing baselines, making pretrained-embedding-based linear-probe workflows feasible for large-scale applications for the first time (Brown et al., 2025).

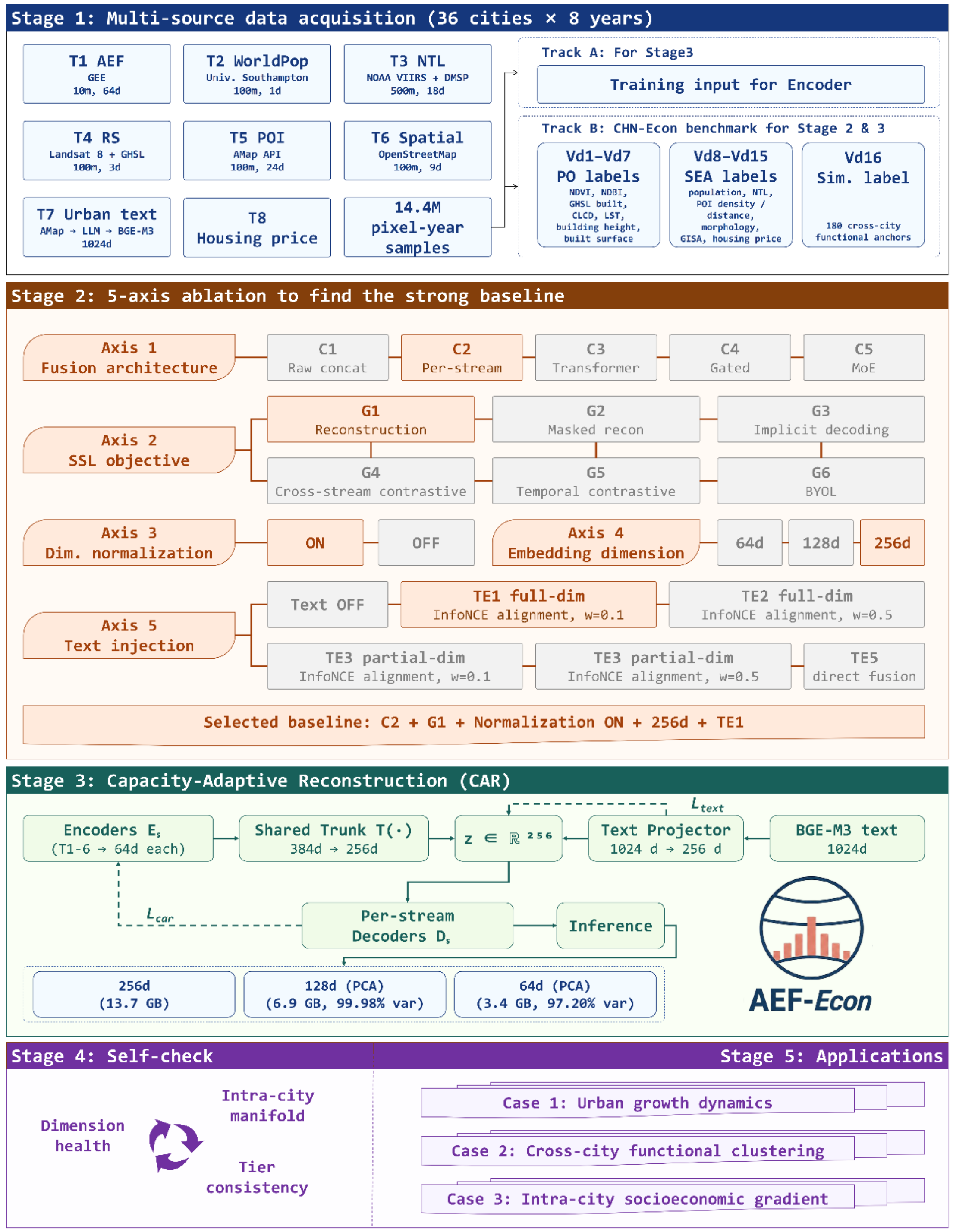

**Fig. 1. Overall technical pipeline of AEF-Econ.**

However, the pretraining objectives of AEF are primarily oriented toward physical land-surface properties, while socioeconomic signals such as population, POIs, and nighttime economic activity are not explicitly incorporated (Hou et al., 2026d; Brown et al., 2025). Several independent evaluations have reported consistent findings across tasks such as gridded population mapping (Hou et al., 2026c), informal settlement identification (Hou et al., 2026e), urban functional zone classification (Liu et al., 2025), poverty mapping (Pettersson and Daoud, 2025), and public health prediction (Nazir et al., 2026): when a task depends on socioeconomic semantics, AEF embeddings require auxiliary signals to reach a usable level of performance. This limitation conflicts with the expected "plug-and-play" role of foundation embeddings.

To address these issues, this study proposes a multi-source self-supervised framework that uses AEF embeddings as the foundation and integrates six types of socioeconomic signals: population, nighttime lights (NTL), remote sensing indices, POIs, urban morphology, and cross-lingual text. Through systematic empirical diagnosis, we identify the bottlenecks in multi-source fusion and introduce a Capacity-Adaptive Reconstruction (CAR) mechanism. The resulting product, AEF-Econ, is designed as a plug-and-play socioeconomic remote sensing foundation embedding. The overall technical workflow is shown in Fig. 1. The main contributions are threefold. First, we integrate seven heterogeneous data streams and construct CHN-Econ, a socioeconomic benchmark covering 36 cities over eight years with 16 validation labels in three categories. Based on 31 controlled experiments, we systematically quantify both the adaptation gap of AEF and the inter-stream capacity competition bottleneck in multi-source fusion. Second, we propose the CAR framework, which extends capacity-fair constraints to the output side through per-stream independent decoding and stream-level reconstruction losses. The full framework improves cross-region $R^2$ from 0.301 using AEF alone to 0.832 with the five-axis backbone and further to 0.848 with CAR; cross-tier $R^2$ increases from 0.160 to 0.671 and then to 0.693. Third, we release the CAR-based AEF-Econ socioeconomic remote sensing embedding product and accompanying code, and validate its downstream usability through a three-level product self-diagnostic assessment and three urban analysis case studies.

The remainder of this paper is organized as follows. Section 2 reviews related work. Section 3 describes the data and evaluation protocol. Section 4 presents the empirical diagnosis based on 31 controlled experiments. Section 5 introduces the CAR framework and its end-to-end results. Section 6 presents the production of the AEF-Econ product and its application cases. Section 7 discusses limitations and future work, and Section 8 concludes the paper.

## 2. Related Work

### 2.1. Geospatial Foundation Models

Geospatial foundation models have rapidly evolved toward task-agnostic and plug-and-play use, yet input modality, training mechanism, and product maturity have rarely been achieved simultaneously (Hou et al., 2026d). MOSAIKS (Rolf et al., 2021) replaces trainable encoders with random projections, limiting the ability of representation capacity to scale with data volume. SatMAE (Cong et al., 2022), Prithvi (Jakubik et al., 2023), and Scale-MAE (Reed et al., 2022) introduce masked autoencoders but are restricted to optical single-modality inputs. SatCLIP (Klemmer et al., 2023) aligns coordinates with imagery, producing location priors rather than time-varying observations. DOFA (Xiong et al., 2024) unifies multiple sensors but structurally excludes non-imagery sources, whereas Tessera (Feng et al., 2025) is limited to Sentinel-1/2 time series. Moreover, most of these models remain at the research-code stage. AEF (Brown et al., 2025) is the first to integrate nine categories of physical sources with text and releases global annual 10-m embeddings for 2017–2024 through Google Earth Engine (GEE). It reduces average error by approximately 23.9% across 15 cross-domain tasks, leading simultaneously in all three dimensions. However, its training objectives remain centered on physical variables, leaving socioeconomic semantics unmodeled. This product maturity provides an off-the-shelf embedding base for this study, while the semantic gap creates methodological room for improvement (Brown et al., 2025; Hou et al., 2026d).

### 2.2. Multi-Source Self-Supervised Representation Learning

The representation quality of multi-source self-supervised learning (SSL) depends on two key design choices: how heterogeneous data streams are integrated into a shared representation space, and how training signals are extracted from unlabeled data.

At the fusion level, direct concatenation is simple and introduces no additional parameters. However, it allocates capacity according to the original feature dimensionality, allowing high-dimensional streams, such as 64-dimensional AEF embeddings, to dominate the gradients and potentially overwhelm low-dimensional signals. Gated fusion (Hu et al., 2018) introduces learnable scalar gates to adjust the overall contribution of each stream, but its granularity remains at the stream level and cannot selectively preserve information at the dimensional level. Cross-attention (Vaswani et al., 2017) enables sample-wise dynamic fusion and offers the strongest expressive capacity. However, its inter-stream interaction patterns may become unstable under distribution shifts, and it does not impose a global constraint on how much capacity each stream should occupy. Mixture-of-Experts (MoE) models (Shazeer et al., 2017; Fedus et al., 2022) dynamically allocate computational resources through routers and are well suited to heterogeneous inputs. Nevertheless, their routing granularity concerns which expert processes a sample rather than how many dimensions are assigned to each stream, making them only indirectly aligned with the capacity-allocation needs of multi-source fusion.

At the training level, contrastive and non-contrastive learning methods, including SimCLR (Chen et al., 2020), MoCo (He et al., 2019), BYOL (Grill et al., 2020), SimSiam (Chen and He, 2020), and DINO (Caron et al., 2021), have been widely validated in single-modality settings. However, their construction of positive pairs relies on different augmented views of the same source signal, whereas multi-source data streams represent different physical dimensions of the same pixel. Whether these two settings are equivalent remains to be examined. Redundancy-reduction methods, such as Barlow Twins (Zbontar et al., 2021) and VICReg (Bardes et al., 2021), effectively prevent representation collapse through decorrelation, but their constraints operate among dimensions within a single stream and do not address subspace overlap across streams. Cross-modal alignment methods, represented by CLIP (Radford et al., 2021), perform well on large-scale image–text pairs, but they depend on end-to-end dual-encoder training and one-to-one semantic pairing. In our setting, by contrast, AEF embeddings are fixed, and no such pairwise correspondence exists among the multi-source streams.

These approaches each have specific applicability and limitations. However, which fusion–training combination best matches the data structure of multi-source socioeconomic remote sensing remains insufficiently examined on dense multi-source benchmarks. This study first addresses this selection problem through controlled experiments, and then introduces targeted improvements for the limitations revealed by the empirical results, particularly the capacity-allocation problem in which low-dimensional semantic streams are compressed by high-dimensional, strong-signal streams.

### 2.3. AlphaEarth-Based Socioeconomic Research

Since the release of AEF, several independent studies have applied its embeddings to socioeconomic remote sensing tasks and have consistently revealed the same limitation. Hou et al. (2026c) found that AEF must be combined with POIs to achieve usable accuracy in gridded population estimation across 18 cities. Hou et al. (2026e) reported unsatisfactory AEF performance in slum identification across 12 cities. Liu et al. (2025) required an additional multimodal alignment mechanism for urban functional zone classification. Pettersson and Daoud (2025) observed unstable cross-regional generalization in poverty mapping across Sub-Saharan Africa. Nazir et al. (2026) likewise required auxiliary signals to enhance discriminative power in public health prediction. A common pattern emerges: when AEF is used alone as a linear probe, its performance on socioeconomic tasks is far below that observed for physical tasks, forcing researchers to concatenate auxiliary signals manually. However, these remedies are task-specific and not reusable, pointing to an unresolved question: how can socioeconomic signals be uniformly represented within foundation embeddings so that downstream users do not have to reinvent the wheel?

## 3. Data and Benchmark

### 3.1. Study Area

This study adopts a two-stage design consisting of sampled-city validation and nationwide product extrapolation. In the validation stage, we select 36 cities in mainland China. Geographic stratification follows the seven major regional divisions: East China with 10 cities, Central China with 3, South China with 4, North China with 5, Northwest China with 5, Southwest China with 5, and Northeast China with 4. Socioeconomic stratification is based on the 2024 city ranking by Yicai, combined with the classification criteria of the National Bureau of Statistics, and includes T1 super-tier cities (6 cities), T1.5 new first-tier cities (7 cities), T2 second-tier cities (17 cities), T3 third-tier cities (5 cities), and T4 fourth-tier cities (1 city; Lhasa). For each city, the administrative center is used as the reference point, from which a rectangular bounding box is generated by extending along longitude and latitude in WGS84. The physical size of the boxes ranges from 17 × 17 km to 51 × 33 km, with an average of approximately 32 × 31 km. To balance data representativeness and distributional consistency, and to reduce the influence of sparse observations, all bounding boxes are manually checked to ensure that they primarily cover core built-up areas and urban–rural transition zones. The coordinates and stratification attributes of the 36 cities are provided in Appendix A.

### 3.2. Dataset

The data used in this study are divided by purpose into training and validation data. The former, described in Section 3.2.1, comprises seven heterogeneous data streams used to train the AEF-Econ encoder. The latter, described in Section 3.2.2, includes 16 downstream labels used to construct the CHN-Econ dataset and evaluate embedding quality.

#### 3.2.1. Training Data

Each pixel-year sample consists of seven data streams: six numerical streams with a total of 119 dimensions and one text-embedding stream with 1024 dimensions (Chen et al., 2024). The sources and dimensionality of each stream are listed in Table 1, and the production workflow is provided in Appendix B.

**Table 1. Overview of the seven training data streams.**

| ID | Data Stream | Data Source | Data Format | Native Resolution | Data Volume | Dimensionality |
|---|---|---|---|---|---|---|
| T1 | AEF embeddings | GEE (Brown et al., 2025) | Raster | 10 m | 337 GB | 64 |
| T2 | WorldPop population | Univ.Southampton (Lloyd et al., 2019) | Raster | 100 m | 119 MB | 1 |
| T3 | Nighttime light statistics | NOAA VIIRS+DMSP (Elvidge et al., 2017) | Raster | 500 m | 1.6 GB | 18 |
| T4 | Remote sensing indices | Landsat 8+ Global Human Settlement Layer (GHSL) (Corbane et al., 2021) | Raster | 100 m | 34 GB | 3 |
| T5 | POI grids | Amap API | Vector (point) | 100 m | 359 GB; 438 million records | 24 |
| T6 | Urban morphology | OpenStreetMap (Haklay and Weber, 2008) | Vector (point/line/polygon) | 100 m | 998 MB | 9 |
| T7 | POI textual embeddings | Amap → BGE-M3 (Chen et al., 2024) | Vector | 100 m | 27.5 GB | 1024 |

T7 is designed to extract functional-zone semantics not captured by T5, including business composition and service tier, rather than quantity or spatial structure, which are already encoded by the 24-dimensional density and distance features in T5. As shown in Fig. 2, the production workflow first extracts all points of interest (POIs) within a 200-m buffer around each pixel. A large language model (LLM) then rewrites these POIs into de-identified functional-zone descriptions using a structured prompt (Fig. 3) (Hou et al., 2026a; Hou et al., 2025b; Hou et al., 2025a; Hou et al., 2026b), retaining only city tier, location context, functional category, and business tier while excluding brand names and geographic proper names. The descriptions are subsequently encoded by BGE-M3 into 1024-dimensional vectors. Table 2 provides examples of raw inputs and abstracted outputs under three representative density scenarios. The purpose of LLM-based abstraction is to reduce the influence of city-specific tokens on cross-city transfer. For example, although two areas in Shanghai and Wuhan may both represent community commercial zones, their POI names may be mapped by BGE-M3 to distant positions in semantic space. After LLM-based abstraction, functionally similar areas are pulled closer in the embedding space (lower left of Fig. 2). Both raw and abstracted embeddings are comparatively evaluated in Section 4.1.5.

**Table 2. Representative examples of LLM-based POI abstraction under sparse, medium, and dense scenarios.**

| ID | Raw Input | Output |
|---|---|---|
| E1 (Sparse) | [Urban fringe] Nearby 200m POI (very few records): Transportation (1): bus stop only | Low-density urban fringe area with minimal functional facilities. Only basic transportation access is present, with no commercial, residential, or service clusters observed, indicating a non-developed or peripheral zone. |

| ID | Raw Input | Output |
|---|---|---|
| E2 (Medium) | [Mid-sized city] Nearby 200m POI (23 records): Shopping (10): local stores, small supermarkets; Food & Beverage (5): local restaurants, fast food; Services (4): courier points, printing shops; Transport (2): bus stops; Accommodation (2): small inns | Medium-density mixed-use urban area dominated by community-level retail and food services. Basic logistics and transportation services are present, with limited accommodation facilities, reflecting a typical residential-commercial neighborhood. |
| E3 (Dense) | [Hangzhou 2022] Nearby 200m POI (95 records): Shopping (35): INTIME In77, HARMAY, Apple Store; Finance (18): CCB private banking center, securities headquarters; Food & Beverage (15): premium restaurants, Starbucks Reserve; Education (10): universities and schools; Accommodation (8): luxury hotels | High-density core urban commercial district in a first-tier city. Dominated by high-end retail and financial services, supported by premium dining, hospitality, and education facilities, exhibiting a highly mixed and fully developed central business district structure. |

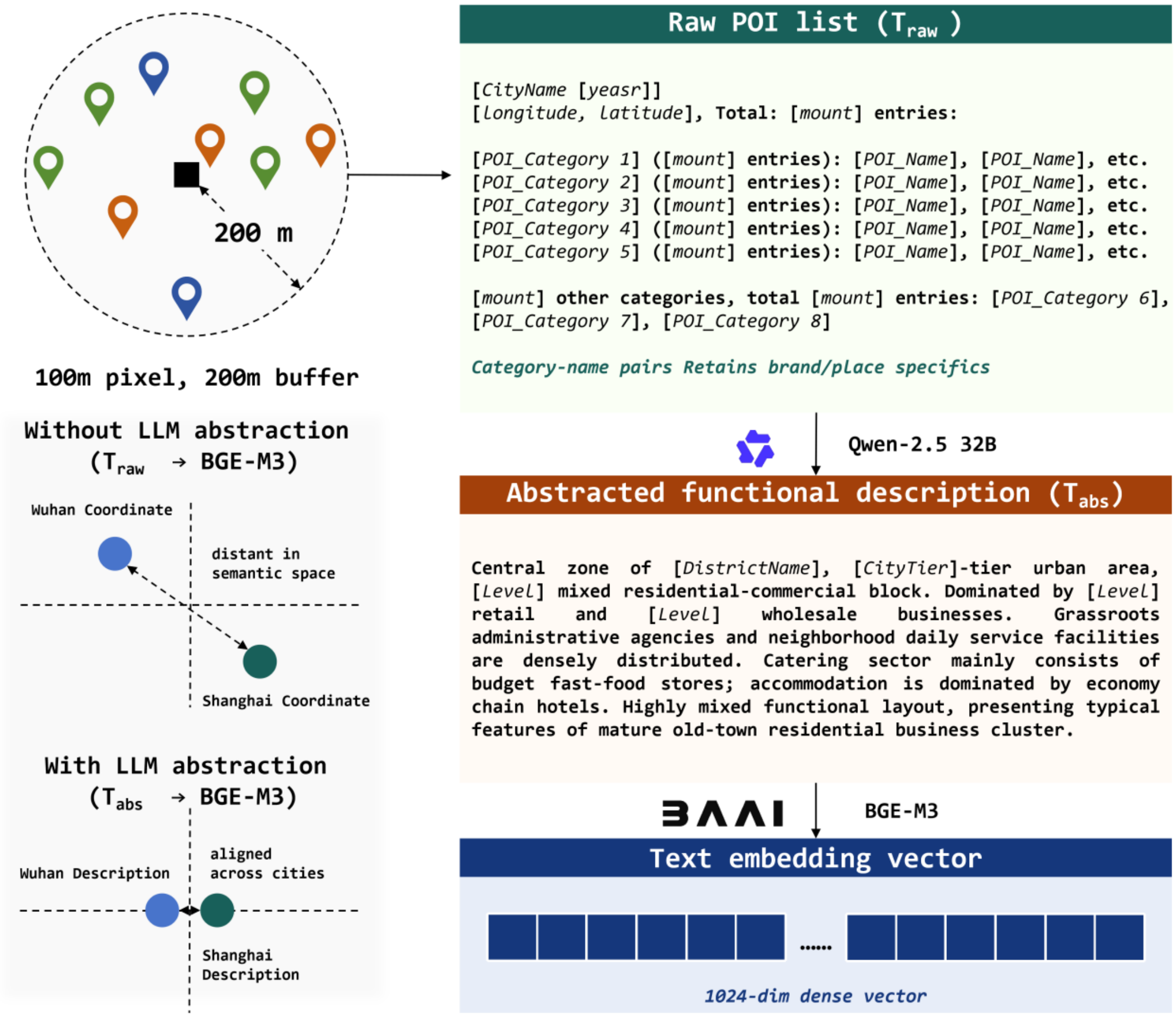

**Fig. 2. T7 text embedding pipeline: from 200 m POI neighborhood extraction through LLM abstraction to BGE-M3 encoding, with cross-city semantic alignment effect (left panels).**

```
Prompt
You are an expert in urban functional zone analysis. Your task is to transform a list of Points of Interest (POIs) within
a 100m × 100m grid cell into a concise functional zone description.Given the POI information of a grid cell, generate a
compact paragraph that characterizes its urban functional profile based on POI category composition, density, business
level, and spatial cues inferred from POI names.

## Requirements:

1. Strictly do NOT include any specific identifiers, including brand names, store names, full institution names,
residential community names, street names, or coordinates.

2. The output must follow four components in strict order (all required):

    ### (a) Spatial location:
    Infer the administrative district and its intra-city spatial position (e.g., central, northern, southern part of a
    district) based on POI cues.
    Format: "{city_tier} city + district + spatial orientation"

    ### (b) Functional composition:
    Identify dominant and secondary urban functions using ONLY the following categories:
    Commercial retail, Catering services, Residential community, Primary education, Higher education, Healthcare services,
    Government administration, Industrial and office, Transportation and logistics, Cultural and leisure, Tourism services,
    Financial services, Religious facilities

    ### (c) Business level (socioeconomic tier):
    Classify the overall level into one of: high-end, mid-to-high-end, medium, mass-market, basic.

    Criteria:
    High-end: international flagship stores, five-star hotels, private international schools, private banking
    Mid-to-high-end: well-known chain brands, boutique hotels, high-quality private schools
    Medium: standard chain stores, budget hotels, public schools
    Mass-market: community shops, low-cost restaurants, small inns
    Basic: wholesale shops, street vendors, short-term rentals

    ### (d) Density and functional mixing:
    POI count > 80 = high density
    20–80 = medium density
    5–20 = low density
    < 5 = very low density
    Category diversity ≥ 5 = high mix
    3–4 = medium mix
    ≤ 2 = low mix

3. The output must be 80–120 words.

4. Output only the final paragraph. Do not include headings, lists, explanations, or extra text.

Input:
{poi_text}
```

**Fig. 3. Structured LLM prompt for de-identified functional zone description, enforcing four mandatory components: spatial location, functional composition, business level, and density characterization.**

#### 3.2.2. Validation Data

The 16 validation labels are divided into two groups according to signal type. The Physical-Observable (PO) group (Vd1–Vd7) covers physical representations explicitly included in AEF pretraining and is used to test whether physical-task performance is preserved after incorporating human-settlement data streams. The Socioeconomic-Anchored (SEA) group (Vd8–Vd16) depends on human-settlement semantics and is used to evaluate gains in the socioeconomic dimension. Some SEA labels share sources with the input streams, such as Vd8 with T2 and Vd9 with T3. This setting is intended to test knowledge internalization rather than constitute data leakage. The data sources and task types of all labels are listed in Table 3, and the production workflow is provided in Appendix C.

**Table 3. CHN-Econ benchmark: 16 validation labels, sources, and task types.**

| Category | ID | Label | Source | Task |
|---|---|---|---|---|
| PO | Vd1–Vd3 | Normalized Difference Vegetation Index (NDVI)/ Normalized Difference Built-up Index (NDBI) /GHSL built-up fraction | Landsat 8/GHSL (Corbane et al., 2021) | Regression |
| | Vd4 | China Land Cover Dataset (CLCD) land use | Wuhan University CLCD (Yang and Huang, 2021) | Classification |
| | Vd5–Vd7 | Land surface temperature/building height/building footprint area | MODIS/GHS (Wan et al., 2015; Corbane et al., 2021) | Regression |

| Category | ID | Label | Source | Task |
|---|---|---|---|---|
| SEA | Vd8–Vd13 | Population/(NTL)/POI/urban morphology | Same as T2–T6 | Regression |
| | Vd14 | First built-up year of impervious surface | Wuhan University Global Impervious Surface Area (GISA) (Huang et al., 2022) | Regression |
| | Vd15 | Intra-urban housing price | Average listing price | Regression |
| | Vd16 | Cross-city functional-zone anchors | 180 manually annotated anchors | Similarity |

### 3.2.3. Data Scale

For each city and year, 50,000 pixels are uniformly sampled using a fixed random seed (seed=42), yielding 36 × 8 × 50,000=14.4 million pixel-year samples in total. All samples are used for unlabeled self-supervised pretraining. For downstream evaluation, the training and test sets are constructed according to the three split protocols described in Section 3.3.2.

## 3.3. Evaluation

Sections 4 and 5 use the same evaluation protocol, which is defined in this section.

### 3.3.1. Evaluation Metrics

The 16 downstream tasks are quantitatively evaluated using different metrics according to task type (Table 4).

**Table 4. Evaluation metrics by task type.**

| Task Type | Labels | Metrics |
|---|---|---|
| Regression | Vd1–Vd3,Vd5–Vd15 | $R^2$ for variance explained and Spearman's $\rho$ for rank preservation |
| Classification | Vd4 | Macro-F1 and $Cohen's\ \kappa$ |
| Similarity | Vd16 | Recall@8+mean reciprocal rank (MRR) |

Overall $R^2$ is computed as the simple average of the lower-bound-clipped $R^2$ values across the 14 regression labels, including Vd1–Vd3 and Vd5–Vd15:

$$\text{Overall } R^2 = \frac{1}{14}\sum_{v} \max\left(R_v^2, -1\right) \quad (1)$$

This is verified as follows:

The theoretical upper bound of $R^2$ is 1, whereas its lower bound is unbounded. In cross-domain settings, individual labels may yield extreme negative values far below −1, such as −14.439 for Vd15 under TEXT_OFF E3. Clipping the lower bound at −1 prevents a single catastrophic label from dominating the overall score, while still retaining moderate negative values to distinguish performance that is slightly worse than the mean baseline from complete failure. This clipping is applied only to the overall aggregation; all label-level results reported throughout the paper use the original, unclipped values. Overall $\rho$ is computed as the simple average of Spearman's $\rho$ across the same 14 labels, without clipping. Vd4, a classification task evaluated by Macro-F1 and $Cohen's\ \kappa$, and Vd16, a retrieval task evaluated by Recall@8 and MRR, are excluded from the overall metrics because their metric scales differ from those of the regression tasks. Their results are reported separately for each experiment in Sections 4 and 5.

### 3.3.2. Evaluation Strategy

Spatial autocorrelation causes within-city random splits to systematically overestimate generalization ability (Tobler, 1970; Roberts et al., 2017). We therefore design three progressively stricter split protocols. E1, within-city random split, uses a 70/30 pixel-level split across the 36 cities and serves as an upper-bound reference for performance. E2, cross-region split, uses 26 cities from East China, Central China, South China, North China, and Northeast China for training, and 10 cities from Northwest and Southwest China for testing. Because these two sets differ systematically in climate, terrain, and economic development, E2 provides a stringent test of cross-regional generalization. E3, cross-tier split, adopts five-fold leave-one-tier-out cross-validation, in which T1, T1.5, T2, T3, and T4 cities are each held out in turn; the main text reports the mean over the five folds. The three split protocols are applied independently to Vd1–Vd15, whereas Vd16 is evaluated through full-database retrieval.

#### 3.3.3. Evaluation Method

Embedding quality is evaluated using linear probes. Regression tasks are evaluated with RidgeCV, with $\alpha \in \{10^{-2},...,10^{2}\}$; classification tasks use L2-regularized logistic regression (Pedregosa et al., 2012); and the similarity task is evaluated by retrieval based directly on cosine similarity. To control computational cost, the training samples for regression tasks are subsampled to 200,000. The use of linear probes follows the principle that embedding quality should be revealed by the simplest possible classifier (Klemmer et al., 2023; Brown et al., 2025). It is also consistent with the plug-and-play and ease-of-use positioning of current remote sensing embedding products: their goal is to allow users to exploit embedding representations through simple downstream models, without deep-learning hyperparameter tuning or model training.

## 4. Empirical Diagnosis

This section has two objectives. First, it searches for the best combination among existing multi-source SSL methods to establish the performance upper bound of available tools. Second, it identifies unresolved issues through controlled experiments, providing empirical evidence for the design of CAR in Section 5. This section includes 31 controlled experiments in total, consisting of five groups of naive baselines (nine experiments in total, with BL2 containing five single-source variants) and 22 methodological ablation experiments.

### 4.1. Experimental Strategy

The experiments are organized into three levels. We first establish naive baselines as reference points, and then search for the optimal configuration along five methodological axes: fusion architecture, self-supervised objective, normalization, embedding dimensionality, and text integration. Each axis is optimized based on the best option identified in the preceding axis, and the process ultimately converges to a fixed backbone configuration.

#### 4.1.1. Naive Baselines

The five groups of baselines approximate the performance boundary from different perspectives (Table 5).

**Table 5. Naive baselines: input configuration and purpose.**

| ID | Baseline | Input | Encoding | Purpose |
|---|---|---|---|---|
| BL1 | AEF only | AEF 64d | Pre-encoded | Stand-alone capacity of AEF |
| BL2 | Single-source | Each individual source | None | Discriminative power of each single stream |
| BL3 | Raw all | All 119d features | None | Upper bound of raw full features |
| BL4 | Raw w/o AEF | 55d features | None | Marginal contribution of AEF |
| BL5 | Random encoder | All 119d features | Random C2 → 256d | Random-encoding reference |

#### 4.1.2. Fusion Architecture

The fusion architecture determines how the six numerical streams, with 119 dimensions in total, are integrated into a d-dimensional dense representation, z. In this section, we fix $d = 256$ to provide sufficient capacity for the encoder. If the model is constrained by a dimensional bottleneck from the outset, the relative differences among architectures may be obscured; the compressibility of the embedding dimensionality is therefore examined in Section 4.1.4. The training objective is temporarily set to G1 reconstruction, which is systematically compared in Section 4.1.3. On this basis, we evaluate five architectures one by one.

C1, direct concatenation, concatenates the six data streams into a 119-dimensional vector, which is then encoded by a shared multilayer perceptron (MLP) into a d-dimensional representation. C2, concatenation after independent projection, first maps each data stream into a 64-dimensional space through a stream-specific MLP. The six projected outputs are then concatenated into a 384-dimensional vector and compressed by a shared MLP into a d-dimensional representation. C3, Transformer-based fusion, treats each projected data stream as a token and models inter-stream interactions using a two-layer, four-head self-attention module, followed by mean pooling and projection into a d-dimensional representation. C4, gated weighted fusion, uses a gating network to generate softmax weights for the six data streams from the full 119-dimensional input. The streams are then combined by weighted summation and compressed by a shared MLP into a d-dimensional representation. C5, MoE fusion, uses a router to select k=2 experts from four expert networks for each sample, and the final output is computed as a weighted combination of the selected expert outputs.

The formulations and network details of each architecture are provided in Appendix D.2. The five architectures mainly differ in their information-fusion mechanisms and in whether the fusion process can dynamically adjust the contribution of each data stream according to the input sample (Table 6). Here, "dynamic adjustment" means that different samples may receive different stream weights, inter-stream interaction patterns, or expert combinations; otherwise, all samples share the same fusion rule.

**Table 6. Comparison of five fusion architectures.**

| ID | Fusion Mechanism | Learned Fusion Weights | Input-Adaptive Adjustment |
|---|---|---|---|
| C1 | Direct concatenation | No | No |
| C2 | Concatenation after independent projection | Yes | No |
| C3 | Attention mechanism | Yes | Yes |
| C4 | Gated weighting | Yes | Yes |
| C5 | Expert routing | Yes | Yes |

#### 4.1.3. Self-Supervised Training Objectives

Self-supervised objectives determine what training signals the model extracts from unlabeled data. G1–G3 are reconstruction-based objectives, where the learning signal is derived from reconstructing the input data. G4–G6 are contrastive-learning and self-distillation objectives, where the learning signal comes from modeling sample-wise similarity relationships. Based on the optimal fusion architecture identified in Section 4.1.2, this section systematically compares the performance of six training objectives.

**G1 (reconstruction):** The encoder compresses the 119-dimensional input into a d-dimensional bottleneck representation, z, and a shared decoder reconstructs the full input. This objective does not rely on data augmentation or negative samples, making it the training paradigm with the weakest assumptions.

$$\hat{x}_s = \mathrm{Dec}_s(z), \quad \mathcal{L}_{G1} = \frac{1}{S}\sum_{s=1}^{S} \| \hat{x}_s - x_s \|^2 \tag{2}$$

**G2 (masked reconstruction):** Each stream is independently masked with a probability of 30%, and the masked streams are reconstructed from the remaining streams. This objective evaluates the encoder's ability to capture statistical dependencies among streams.

$$\mathcal{L}_{G2} = \sum_{s \in M} \| \hat{x}_s - \mathrm{Dec}_s(\mathrm{Enc}(x_{\setminus M})) \|^2 \tag{3}$$

**G3 (implicit decoding):** At each step, the decoder receives only z and the index of the queried dimension, and predicts the value of that dimension. This objective tests the dimension-wise accessibility of information encoded in the embedding.

$$\mathcal{L}_{G3} = \frac{1}{n_q}\sum_{k=1}^{n_q} \| f_\theta(z, e_{i_k}) - x_{i_k} \|^2 \tag{4}$$

**G4 (cross-stream contrast):** Two streams are randomly sampled and encoded as a positive pair. An noise contrastive estimation (InfoNCE) loss with $\tau = 0.1$ is used to pull representations of the same pixel closer while pushing apart those of different pixels. This objective tests the spatial co-location assumption.

$$\mathcal{L}_{G4} = -\frac{1}{B}\sum_{i=1}^{B} \log \frac{\exp\left(\frac{\mathrm{sim}\left(z_i^a, z_i^b\right)}{\tau}\right)}{\sum_{j=1}^{B} \exp\left(\frac{\mathrm{sim}\left(z_i^a, z_j^b\right)}{\tau}\right)} \tag{5}$$

**G5 (temporal contrast):** The positive pair is defined as the representations of the same pixel in adjacent years, testing the temporal proximity assumption.

$$\mathcal{L}_{G5} = -\frac{1}{B}\sum_{i=1}^{B} \log \frac{\exp(\mathrm{sim}(z_{p_i,t}, z_{p_i,t+1})/\tau)}{\sum_{j=1}^{B} \exp\left(\mathrm{sim}(z_{p_i,t}, z_{p_j,t+1})/\tau\right)} \tag{6}$$

**G6 (BYOL):** Different subsets of the multi-source streams are treated as two views. An asymmetric online–target network architecture with momentum updating (m=0.99) is used to prevent representation collapse.

$$\mathcal{L}_{G6} = 2 - 2 \cdot \frac{\langle p, z' \rangle}{\|p\| \cdot \|z'\|} \tag{7}$$

The loss formulations and hyperparameters for each objective are provided in Appendix D.3.

#### 4.1.4. Normalization and Embedding Dimensionality

After determining the fusion architecture and self-supervised objective, two additional engineering choices that affect the product specification require empirical validation.

Normalization strategy: The six numerical streams differ substantially in physical scale, ranging from NTL radiance values of 0–1000, to normalized NDVI values of −1 to 1, and POI counts from 0 to several thousand. Under the G1 reconstruction objective, the reconstruction errors of large-scale streams are numerically much larger than those of small-scale streams, causing gradients to be dominated by high-magnitude streams. per_dim_norm applies within-batch variance normalization independently to each output dimension of the reconstruction loss, rescaling dimension-wise errors to comparable magnitudes. Based on the optimal fusion architecture and training objective identified in the preceding two sections, this section compares two configurations: with and without per_dim_norm.

Embedding dimensionality: Section 4.1.2 uses d=256 as the default search setting. However, if 128 or even 64 dimensions can preserve accuracy, the product storage requirement can be reduced to one quarter. After determining the normalization strategy, this section compares the accuracy degradation across three embedding dimensions: 64, 128, and 256.

#### 4.1.5. Text-Modality Integration

The text stream T7, represented by 1024-dimensional BGE-M3 embeddings, is not included in the naive baselines. The 1024-dimensional semantic vectors and the 119-dimensional numerical streams lie on different manifolds; directly concatenating them as input to Ridge would allow the text stream to dominate the numerical streams through an 8.6-fold dimensional advantage. In addition, BGE-M3 outputs have no physical raw values and are therefore unsuitable for the mean squared error (MSE) reconstruction objective used in G1. Text integration thus requires a dedicated design. Building on the numerical backbone obtained through the axis-wise search in Sections 4.1.2–4.1.4, this section evaluates three types of text-integration strategies comprising five configurations, TE1–TE5.

TE1/TE2 (full-dimensional InfoNCE alignment): A text projection head (1024 → 512 → 256) is added, and bidirectional InfoNCE with $\tau = 0.1$ is used to pull the numerical representation z and the text representation t of the same pixel closer. The total loss is defined as $\mathcal{L} = \mathcal{L}_{\mathrm{G1}} + \mathrm{w} \cdot \mathcal{L}_{\mathrm{text}}$. The forward computation of z does not depend on text; instead, text guides z toward the semantic direction only through gradients. Therefore, no text input is required during inference. TE1 uses w=0.5, and TE2 uses w=0.1.

$$\mathcal{L}_{text} = \frac{1}{2}[InfoNCE(z \to t) + InfoNCE(t \to z)], \quad \tau = 0.1 \tag{8}$$

TE3/TE4 (partial-dimensional InfoNCE alignment): InfoNCE is restricted to the last 64 dimensions of z, while the first 192 dimensions are driven only by $\mathcal{L}_{\mathrm{G1}}$ and are completely isolated from text gradients. The output of the text projection head is reduced to 64 dimensions. TE3 uses w=0.5, and TE4 uses w=0.1.

TE5 (direct text fusion): The alignment paradigm is abandoned. Instead, the text projection (1024 → 64) is directly concatenated with the projections of the six numerical streams to form a 448-dimensional vector, which is then compressed by the trunk to *d*=256. The learning objective remains $\mathcal{L}_{\mathrm{G1}}$. Unlike TE1–TE4, TE5 still requires text input during inference.

### 4.2. Experimental Results

#### 4.2.1. Results of BL1–BL5

The overall performance of the nine baselines, including BL1, the five single-source variants of BL2, BL3, BL4, and BL5, is reported in Table 7.

**Table 7. Overall performance ($R^2$ and Spearman ρ) of naive baselines under three split protocols.**

| Strategy | E1 | | E2 | | E3 | |
|---|---|---|---|---|---|---|
| | $R^2$ | ρ | $R^2$ | ρ | $R^2$ | ρ |
| BL1 | 0.635 | 0.789 | 0.301 | 0.628 | 0.160 | 0.692 |
| BL2-WP | 0.163 | 0.300 | 0.068 | 0.293 | −0.285 | 0.230 |
| BL2-NTL | 0.426 | 0.638 | 0.362 | 0.612 | 0.152 | 0.576 |
| BL2-RS | 0.461 | 0.658 | 0.407 | 0.619 | 0.204 | 0.583 |
| BL2-POI | 0.384 | 0.622 | 0.256 | 0.580 | 0.022 | 0.562 |
| BL2-SP | 0.402 | 0.607 | 0.316 | 0.572 | 0.054 | 0.527 |
| BL3 | **0.844** | **0.913** | **0.772** | **0.860** | 0.561 | **0.852** |
| BL4 | 0.799 | 0.886 | 0.745 | 0.850 | **0.606** | 0.830 |
| BL5 | 0.803 | 0.867 | 0.736 | 0.822 | 0.534 | 0.788 |

BL3, which uses all 119 dimensions, achieves the best performance in five of the six evaluation units. In contrast, BL1, which uses AEF alone, and all single-source variants in BL2 lag substantially behind. Even the strongest BL2 variant, BL2-RS, reaches only 0.461 on E1, compared with 0.635 for BL1, indicating that neither AEF nor any individual source alone can sufficiently cover the information requirements of socioeconomic tasks. BL4, which removes AEF and retains the remaining 55 dimensions, yields $R^2$ values only 0.02–0.05 lower than BL3 under the E1 and E2 splits, suggesting that the marginal contribution of AEF is limited. Under E3, BL4 surpasses BL3 in $R^2$, with 0.606 versus 0.561, but its $\rho$ is 0.022 lower. Because $R^2$ can be strongly affected by a small number of extreme predictions, whereas $\rho$ is more robust to outliers, this $R^2$ advantage of BL4 may result from incidental fluctuations in individual labels. BL5 uses the same C2 network architecture as the subsequent SSL encoder but with randomly initialized weights. Its E1 $R^2$ of 0.803 is lower than that of BL3 without an encoder (0.844), indicating that an untrained encoder instead loses information. Therefore, if a trained encoder fails to outperform BL3, the training should be considered ineffective.

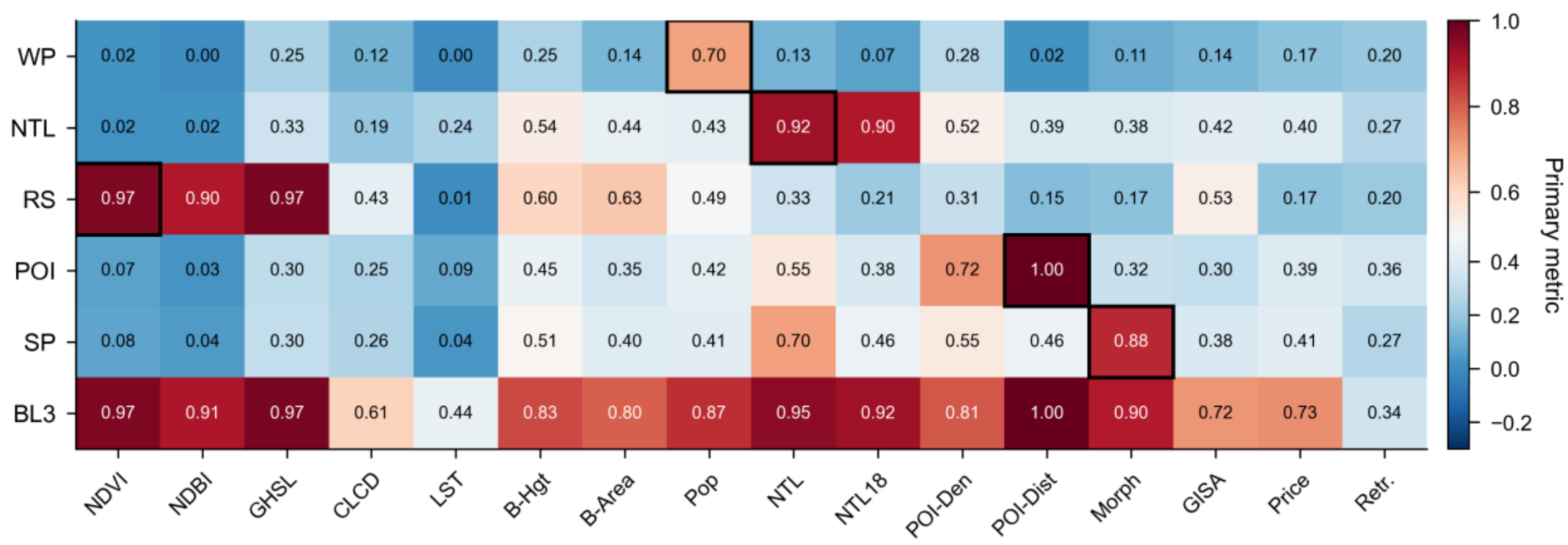

**Fig. 4. Per-label R² heatmap of single-source baselines (BL2) versus the full-feature baseline (BL3), showing complementary information across data streams.**

The label-level heatmap in Fig. 4 reveals the typical behavior of single-source representations. Each data stream shows a clear peak on its source-matched labels, such as RS for NDVI/GHSL ($R^2 \approx 0.97$), NTL ($R^2 \approx 0.92$), POI for POI distance ($R^2 \approx 1.00$), and SP for spatial morphology ($R^2 \approx 0.88$), indicating that source-specific information is effectively preserved. However, these high values are largely confined to a small number of labels, while each single source shows limited predictive capacity for the remaining tasks. This pattern suggests that different data streams capture only partial aspects of socioeconomic phenomena. By contrast, BL3 maintains strong performance across nearly all labels, preserving stream-specific information while integrating cross-source complementary knowledge. Its overall $R^2$ remains 0.561 under E3, whereas the single-source baselines range only from −0.285 to 0.204, indicating that multi-source information is complementary rather than redundant. How to effectively fuse these distributed signals into a unified representation therefore becomes the central question of the subsequent experiments.

**Table 8. Vd4 (CLCD land-use classification) performance of naive baselines.**

| Configuration | E1 F1 | E1κ | E2 F1 | E2κ | E3 F1 | E3κ |
|---|---|---|---|---|---|---|
| BL1 | 0.601 | 0.783 | 0.466 | 0.594 | 0.484 | 0.645 |
| BL2-WP | 0.123 | 0.165 | 0.125 | 0.143 | 0.112 | 0.148 |
| BL2-NTL | 0.191 | 0.339 | 0.119 | 0.183 | 0.137 | 0.207 |
| BL2-RS | 0.427 | 0.597 | 0.256 | 0.337 | 0.317 | 0.423 |
| BL2-POI | 0.253 | 0.371 | 0.133 | 0.211 | 0.178 | 0.233 |
| BL2-SP | 0.262 | 0.369 | 0.148 | 0.223 | 0.188 | 0.227 |
| BL3 | **0.611** | **0.795** | **0.482** | **0.643** | **0.488** | **0.667** |
| BL4 | 0.523 | 0.691 | 0.360 | 0.487 | 0.411 | 0.512 |
| BL5 | 0.589 | 0.769 | 0.380 | 0.568 | 0.441 | 0.576 |

**Table 9. Vd16 (cross-city functional anchor retrieval) performance of naive baselines.**

| Configuration | Recall@8 | MRR |
|---|---|---|
| BL1 | 0.244 | 0.450 |
| BL2-WP | 0.199 | 0.445 |
| BL2-NTL | 0.267 | 0.505 |
| BL2-RS | 0.201 | 0.417 |
| BL2-POI | 0.356 | 0.578 |
| BL2-SP | 0.266 | 0.464 |
| BL3 | **0.361** | **0.588** |
| BL4 | 0.339 | 0.575 |
| BL5 | 0.249 | 0.484 |

In the Vd4 classification task (Table 8) and the Vd16 retrieval task (Table 9), the overall conclusions are consistent with those from the $R^2$-based regression tasks. Both tasks show that multi-source fused representations, represented by BL3 and its variants, consistently outperform models based on individual data streams. BL3 therefore serves as a strict lower bound that subsequent encoders must exceed.

#### 4.2.2. Results of C1–C5

The overall performance of the five fusion architectures is reported in Table 10. The most direct finding is that C2 achieves the best performance across all six regression evaluation units, with $R^2$ values of 0.885, 0.802, and 0.620 under E1, E2, and E3, respectively. More importantly, its advantage increases as the split becomes more challenging: C2 exceeds C5 by only 0.007 under E1, but the margin expands to 0.054 under E3. This suggests that differences among fusion architectures are reflected primarily in generalization ability rather than in fitting capacity within the training distribution.

**Table 10. Overall performance ($R^2$ and Spearman ρ) of five fusion architectures (C1–C5) under three split protocols.**

| Strategy | E1 | | E2 | | E3 | |
|---|---|---|---|---|---|---|
| | $R^2$ | ρ | $R^2$ | ρ | $R^2$ | ρ |
| C1 | 0.856 | 0.912 | 0.695 | 0.836 | 0.423 | 0.843 |
| C2 | **0.885** | **0.923** | **0.802** | **0.868** | **0.620** | **0.861** |
| C3 | 0.831 | 0.898 | 0.770 | 0.852 | 0.490 | 0.840 |
| C4 | 0.793 | 0.877 | 0.716 | 0.827 | 0.410 | 0.805 |
| C5 | 0.878 | 0.916 | 0.793 | 0.852 | 0.566 | 0.845 |

The superior performance of C2 stems from its use of independent projection for each data stream before entering the shared representation space. This design ensures that different data streams are assigned comparable representational capacity, allowing both the one-dimensional WorldPop stream and the 64-dimensional AEF stream to obtain independent representations before fusion, rather than being prematurely overwhelmed by differences in their original dimensionality. For cross-city and cross-regional settings, such capacity balancing is more important than more complex fusion mechanisms.

**Table 11. Vd4 (CLCD land-use classification) performance of five fusion architectures.**

| Configuration | E1 F1 | E1κ | E2 F1 | E2κ | E3 F1 | E3κ |
|---|---|---|---|---|---|---|
| C1 | 0.621 | 0.799 | 0.400 | 0.589 | 0.468 | 0.620 |
| **C2** | **0.624** | **0.820** | **0.487** | **0.684** | **0.499** | **0.684** |
| C3 | 0.589 | 0.786 | 0.429 | 0.659 | 0.464 | 0.622 |
| C4 | 0.600 | 0.783 | 0.438 | 0.677 | 0.466 | 0.629 |
| C5 | 0.623 | 0.800 | 0.413 | 0.609 | 0.469 | 0.630 |

The classification and retrieval results further support this conclusion. In the Vd4 classification task (Table 11), C2 again achieves the best performance, with F1 scores of 0.487 and 0.499 under E2 and E3, respectively, and κ values of 0.684 in both settings, clearly outperforming the other fusion architectures. In the Vd16 retrieval task (Table 12), C2 also obtains the highest Recall@8 (0.372) and MRR (0.608), exceeding more complex structures such as C5 (0.349/0.601). These results indicate that the advantage of C2 is not limited to $R^2$-based regression tasks, but also transfers to classification and similarity-based retrieval tasks.

**Table 12. Vd16 (cross-city functional anchor retrieval) performance of five fusion architectures.**

| Configuration | Recall@8 | MRR |
|---|---|---|
| C1 | 0.336 | 0.585 |
| C2 | **0.372** | **0.608** |
| C3 | 0.326 | 0.552 |
| C4 | 0.302 | 0.525 |
| C5 | 0.349 | 0.601 |

The performance of the remaining architectures further confirms this finding. C1 uses the simplest direct-concatenation strategy and reaches 0.856 under E1, only slightly below C2. However, its performance drops to 0.423 under E3, widening the gap to 0.197. This suggests that although the model can exploit multi-source information, it struggles to learn stable cross-domain fusion rules.

C4 performs worst in five of the six regression evaluation units, with $R^2$ values of 0.793 under E1 and 0.410 under E3. It also obtains the lowest Recall@8 (0.302) and MRR (0.525) on Vd16, indicating that compressing multi-stream information into a single weighted representation leads to substantial information loss. By contrast, although C3 and C5 introduce more complex mechanisms, such as attention and MoE, their E3 $R^2$ values reach only 0.490 and 0.566, respectively, both below C2. They also show no stable advantage on Vd4 or Vd16. These results indicate that the key challenge in the current task is not to model more complex inter-stream relationships, but to ensure that information from each data stream is sufficiently preserved.

Therefore, this set of experiments conveys a clear message: multi-source fusion must first address whether information is fairly preserved, and only then how complex interactions should be modeled. C2 satisfies the former in the most direct way and delivers the most stable performance across regression, classification, and retrieval tasks. It is therefore selected as the fusion-architecture backbone for subsequent experiments.

#### 4.2.3. Results of G1–G6

The performance differences among the six SSL objectives are substantially larger than those among the fusion architectures examined in the previous section. This indicates that, for multi-source economic representation learning, what the model learns is more important than how the sources are fused. The overall comparison of the six SSL objectives, G1–G6, is reported in Table 13.

**Table 13. Overall performance ($R^2$ and Spearman $\rho$) of six SSL objectives (G1–G6) under three split protocols.**

| Strategy | E1 | | E2 | | E3 | |
|---|---|---|---|---|---|---|
| | R² | ρ | R² | ρ | R² | ρ |
| G1 | **0.885** | **0.923** | **0.802** | **0.868** | **0.620** | **0.861** |
| G2 | 0.814 | 0.868 | 0.662 | 0.770 | 0.293 | 0.755 |
| G3 | 0.880 | 0.916 | 0.755 | 0.862 | 0.504 | 0.835 |
| G4 | 0.824 | 0.876 | 0.765 | 0.831 | 0.579 | 0.793 |
| G5 | 0.647 | 0.778 | 0.201 | 0.569 | 0.255 | 0.616 |
| G6 | 0.790 | 0.857 | 0.714 | 0.815 | 0.488 | 0.777 |

The results show that G1, G3, and G4 are the only clearly effective objectives, with all three achieving E1 $R^2$ values above 0.82 and substantially outperforming the remaining objectives. Among them, G1 performs best across all six regression evaluation units, with $R^2$ values of 0.885, 0.802, and 0.620 under E1, E2, and E3, respectively. It also leads consistently in $\rho$, reaching an E3 ρ of 0.861. This advantage extends to the classification and retrieval tasks. On Vd4 (Table 14), G1 achieves the highest E2/E3 F1 scores (0.487/0.499) and $\kappa$ values (both 0.684). On Vd16 (Table 15), G1 also obtains the highest Recall@8 (0.372) and MRR (0.608). These results indicate that the reconstruction objective not only improves regression performance but also produces a more stable general-purpose representation for classification and similarity-based retrieval.

**Table 14. Vd4 (CLCD land-use classification) performance of six SSL objectives.**

| Configuration | E1 F1 | E1κ | E2 F1 | E2κ | E3 F1 | E3κ |
|---|---|---|---|---|---|---|
| G1 | **0.624** | **0.820** | **0.487** | **0.684** | **0.499** | **0.684** |
| G2 | 0.601 | 0.783 | 0.336 | 0.439 | 0.420 | 0.558 |
| G3 | 0.620 | 0.799 | 0.435 | 0.546 | 0.482 | 0.622 |
| G4 | 0.591 | 0.774 | 0.363 | 0.513 | 0.450 | 0.587 |
| G5 | 0.537 | 0.695 | 0.233 | 0.271 | 0.345 | 0.419 |
| G6 | 0.544 | 0.733 | 0.340 | 0.493 | 0.431 | 0.545 |

Although G4 achieves lower $R^2$ values than G1 under E2 and E3 (0.765/0.579), it clearly outperforms most other objectives, indicating that cross-stream contrastive learning can indeed extract stable structural information shared across data sources. However, the $\rho$ values of G4 are lower than those of G1 under all three splits, and G4 shows no advantage on Vd4 or Vd16. Its E3 $\kappa$ is only 0.587, and its Recall@8 and MRR on Vd16 are 0.285 and 0.543, respectively. This suggests that while the contrastive objective strengthens cross-stream consistency, it may discard some stream-specific details, thereby affecting ranking, classification, and retrieval performance.

By contrast, G5 and G6 perform relatively poorly. G5 reaches only 0.201 in E2 $R^2$, only 0.271 in Vd4 E2 $\kappa$, and its Vd16 Recall@8 drops to 0.201. Although G6 performs slightly better than G5, it still lags substantially behind G1, G3, and G4 overall. These results suggest that excessive reliance on local perturbations or a single pretraining objective is insufficient to support the multi-source information integration required by complex socioeconomic labels.

The comprehensive advantage of G1 stems from the reconstruction objective, which requires the model to recover all input dimensions from the embedding. This objective preserves cross-source shared structures while preventing the model from discarding unique information carried by any data stream. For AEF-Econ, compressing multi-source information as completely as possible into a unified representation is better aligned with the intended goal than extracting a single stable common factor. Therefore, G1 is selected as the SSL-objective backbone for subsequent experiments.

**Table 15. Vd16 (cross-city functional anchor retrieval) performance of six SSL objectives.**

| Configuration | Recall@8 | MRR |
|---|---|---|
| G1 | **0.372** | **0.608** |
| G2 | 0.344 | 0.566 |
| G3 | 0.323 | 0.581 |
| G4 | 0.285 | 0.543 |
| G5 | 0.201 | 0.462 |
| G6 | 0.249 | 0.445 |

#### 4.2.4. Embedding Dimensionality and Normalization Strategy

After C2 and G1 are selected, two key questions remain: how the reconstruction loss should balance different data streams, and how many dimensions the final embedding should retain. Because the six numerical streams differ substantially in scale, directly computing MSE would assign greater loss weights to high-magnitude variables such as NTL and POI, while weakening low-magnitude variables such as NDVI. The role of per_dim_norm is to rescale the error of each output dimension to a comparable magnitude, so that training no longer primarily follows variables with large absolute values but instead preserves information across dimensions more evenly.

**Table 16. Effect of per-dimension normalization on overall performance.**

| Configuration | E1 | | E2 | | E3 | |
|---|---|---|---|---|---|---|
| | $R^2$ | ρ | $R^2$ | ρ | $R^2$ | ρ |
| PERNORM_OFF | 0.885 | 0.923 | 0.802 | 0.868 | 0.620 | 0.861 |
| PERNORM_ON | **0.892** | **0.936** | **0.819** | **0.889** | **0.666** | **0.878** |
| Δ (ON−OFF) | 0.007 | 0.013 | 0.017 | 0.021 | 0.046 | 0.017 |

The results are consistent with this expectation. After enabling per_dim_norm, both $R^2$ and $\rho$ improve under all three split protocols. Specifically, E2 $R^2$ increases from 0.802 to 0.819, and E3 $R^2$ increases from 0.620 to 0.666, with corresponding gains in $\rho$. Similar benefits are observed in the classification and retrieval tasks: the E2/E3 $\kappa$ values for Vd4 increase from 0.684/0.684 to 0.702/0.696, while Recall@8 and MRR for Vd16 improve from 0.372/0.608 to 0.388/0.625. These results indicate that dimension-wise normalization not only improves numerical reconstruction in regression tasks but also helps produce more stable representations for classification and retrieval. The normalization ablation results are reported in Table 16; therefore, subsequent experiments use per_dim_norm=ON.

**Table 17. Overall performance at three embedding dimensions with relative storage cost.**

| Strategy | Storage Ratio | E1 | | E2 | | E3 | |
|---|---|---|---|---|---|---|---|
| | | $R^2$ | ρ | $R^2$ | ρ | $R^2$ | ρ |
| DIM64 | 1× | 0.826 | 0.896 | 0.772 | 0.867 | 0.458 | 0.838 |
| DIM128 | 2× | 0.866 | 0.915 | 0.800 | 0.881 | 0.642 | 0.852 |
| DIM256 | 4× | **0.892** | **0.936** | **0.819** | **0.889** | **0.666** | **0.878** |

Building on this setting, we further compare different embedding dimensionalities (Table 17). The 256-dimensional embedding achieves the highest $R^2$ under all three regression splits, with values of 0.892, 0.819, and 0.666 for E1, E2, and E3, respectively. When compressed to 128 dimensions, $R^2$ decreases by 0.026, 0.019, and 0.024 under E1, E2, and E3, respectively, indicating a relatively modest cost. However, further compression to 64 dimensions reduces E3 $R^2$ to 0.458, suggesting substantial information loss. The same trend is observed for Vd4 (Table 18) and Vd16 (Table 19). DIM256 achieves the highest E2/E3 $\kappa$ values on Vd4 (0.702/0.696), as well as the highest Recall@8 (0.388) and MRR (0.625) on Vd16. DIM128 remains close overall but still underperforms the 256-dimensional setting, whereas DIM64 shows the most pronounced degradation.

**Table 18. Vd4 and normalization/dimension ablation.**

| Configuration | E1 F1 | E1κ | E2 F1 | E2κ | E3 F1 | E3κ |
|---|---|---|---|---|---|---|
| PERNORM_OFF | 0.624 | 0.82 | 0.487 | 0.684 | 0.499 | 0.684 |
| PERNORM_ON | **0.638** | **0.841** | **0.492** | **0.702** | **0.516** | **0.696** |
| DIM64 | 0.592 | 0.785 | 0.418 | 0.637 | 0.474 | 0.651 |
| DIM128 | 0.632 | 0.827 | 0.478 | 0.666 | 0.497 | 0.684 |
| DIM256 | **0.638** | **0.841** | **0.492** | **0.702** | **0.516** | **0.696** |

Based on these results, all subsequent experiments use the independent-projection fusion architecture (C2), the reconstruction objective (G1), dimension-wise normalization enabled (per_dim_norm=ON), and a 256-dimensional embedding configuration.

**Table 19. Vd16 and normalization/dimension ablation.**

| Configuration | Recall@8 | MRR |
|---|---|---|
| PERNORM_OFF | 0.372 | 0.608 |
| PERNORM_ON | **0.388** | **0.625** |
| DIM64 | 0.334 | 0.567 |
| DIM128 | 0.369 | 0.607 |
| DIM256 | **0.388** | **0.625** |

#### 4.2.5. Text-Modality Integration

The overall performance of the six configurations is reported in Table 20. Compared with the preceding experiments, the numerical backbone has already been systematically optimized in terms of fusion architecture, training objective, normalization strategy, and embedding dimensionality. As a result, TEXT_OFF already achieves a strong baseline performance, with an E3 $R^2$ of 0.666. Against this optimized baseline, the room for further improvement from text integration is substantially reduced, but text still provides consistent gains.

**Table 20. Overall performance of text injection configurations under three split protocols.**

| Strategy | E1 | | E2 | | E3 | |
|---|---|---|---|---|---|---|
| | $R^2$ | ρ | $R^2$ | ρ | $R^2$ | ρ |
| TEXT_OFF | 0.892 | 0.936 | 0.819 | 0.889 | 0.666 | 0.878 |
| TE1 (w=0.5) | 0.863 | 0.905 | 0.707 | 0.819 | 0.477 | 0.806 |
| TE2 (w=0.1) | **0.896** | **0.942** | **0.832** | **0.894** | **0.671** | **0.894** |
| TE3 (w=0.5) | 0.861 | 0.904 | 0.654 | 0.817 | 0.517 | 0.802 |
| TE4 (w=0.1) | 0.876 | 0.912 | 0.731 | 0.838 | 0.570 | 0.823 |
| TE5 | 0.876 | 0.912 | 0.682 | 0.856 | 0.548 | 0.830 |

The best-performing configuration, TE2, outperforms TEXT_OFF under all three regression splits. Specifically, E2 $R^2$ increases from 0.819 to 0.832, and E3 $R^2$ increases from 0.666 to 0.671, with corresponding improvements in ρ. The classification and retrieval results show the same trend. On Vd4 (Table 21), TE2 achieves the highest E2/E3 κ values of 0.710 and 0.704, respectively. On Vd16 (Table 22), TE2 reaches a Recall@8 of 0.399 and an MRR of 0.642, again outperforming TEXT_OFF. These results indicate that text does not replace numerical streams; rather, it provides an additional semantic constraint on top of the existing numerical representation.

**Table 21. Vd4 (CLCD land-use classification) performance of text injection configurations.**

| Configuration | E1 F1 | E1κ | E2 F1 | E2κ | E3 F1 | E3κ |
|---|---|---|---|---|---|---|
| TEXT_OFF | 0.638 | 0.841 | 0.492 | 0.702 | 0.516 | 0.696 |
| TE1 (w=0.5) | 0.597 | 0.792 | 0.451 | 0.593 | 0.486 | 0.631 |
| TE2 (w=0.1) | **0.642** | **0.861** | **0.514** | **0.710** | **0.521** | **0.704** |
| TE3 (w=0.5) | 0.611 | 0.792 | 0.334 | 0.443 | 0.391 | 0.497 |
| TE4 (w=0.1) | 0.617 | 0.798 | 0.359 | 0.490 | 0.453 | 0.575 |
| TE5 | 0.624 | 0.803 | 0.378 | 0.541 | 0.478 | 0.622 |

Because the text is derived from semantic abstraction of POI-based functional zones, its contribution is more likely to come from high-level semantics that are difficult for numerical streams to express directly, such as functional type, business structure, and development tier. A consistent pattern also emerges across text-integration strategies: regardless of whether full-dimensional or partial-dimensional alignment is used, the low-weight setting (w=0.1) clearly outperforms the high-weight setting (w=0.5), with TE2 outperforming TE1 and TE4 outperforming TE3. When the text weight is too high, the model overfits to textual semantics and instead weakens the information carried by the numerical streams themselves.

**Table 22. Vd16 performance of text injection configurations.**

| Configuration | Recall@8 | MRR |
|---|---|---|
| TEXT_OFF | **0.388** | **0.625** |
| TE1 (w=0.5) | 0.385 | 0.594 |
| TE2 (w=0.1) | 0.399 | 0.642 |
| TE3 (w=0.5) | 0.333 | 0.574 |
| TE4 (w=0.1) | 0.335 | 0.585 |
| TE5 | 0.321 | 0.569 |

The difference between full-dimensional alignment and alignment restricted to the last 64 dimensions is also clear. TE2 reaches an E3 $R^2$ of 0.671, whereas TE4 reaches only 0.570; TE2 also substantially outperforms TE4 on Vd4 and Vd16. This indicates that applying textual semantic constraints to the entire representation space is more effective than constraining only the last 64 dimensions. By contrast, TE5, which directly fuses text, does not further improve performance. Its regression results under E2 and E3, classification results on Vd4, and retrieval results on Vd16 are all lower than those of TE2. This suggests that text is better suited as an auxiliary constraint during training than as an additional input stream for the current task.

**Table 23. Vd15 metrics tracking across text injection configurations, highlighting cross-domain collapse and recovery.**

| Strategy | E1 | | E2 | | E3 | |
|---|---|---|---|---|---|---|
| | $R^2$ | ρ | $R^2$ | ρ | $R^2$ | ρ |
| TEXT_OFF | 0.763 | 0.872 | −1.471 | 0.596 | −14.439 | 0.499 |
| TE1 (w=0.5) | 0.863 | 0.923 | −0.177 | 0.568 | −0.108 | 0.579 |
| TE2 (w=0.1) | **0.876** | **0.930** | **0.278** | **0.718** | **0.007** | **0.621** |
| TE3 (w=0.5) | 0.875 | 0.930 | −0.798 | 0.556 | 0.035 | 0.563 |
| TE4 (w=0.1) | 0.871 | 0.928 | −0.195 | 0.591 | −0.034 | 0.646 |
| TE5 | 0.813 | 0.903 | −1.304 | 0.543 | −0.140 | 0.578 |

Vd15, the housing-price task (Table 23), further reveals where the value of text comes from. Compared with labels such as NTL, population, or building height, housing price depends more strongly on high-level semantic information, including regional functional positioning, commercial tier, and stage of urban development. It is therefore also the most difficult task for cross-domain generalization. Under TEXT_OFF, E3 $R^2$ drops to −14.439, indicating near-complete model failure. After text is introduced, all configurations show substantial improvement. For example, TE2 raises E3 $R^2$ to 0.007, and TE3 reaches 0.035; even the weakest text configuration, TE5, recovers from −14.439 to −0.140. This demonstrates that text indeed provides cross-city semantic information that numerical streams struggle to express, enabling the model to recognize functional similarity among "same-type areas" rather than relying solely on local numerical features.

Nevertheless, this improvement remains insufficient for reliable housing-price prediction. Although text substantially mitigates cross-domain degradation, the E3 $R^2$ values of all configurations still hover around zero and remain far below those of other labels. This indicates that text addresses the problem of failing to recognize similar functional areas across cities, but it cannot establish a stable cross-city price-mapping relationship. Factors on which housing prices depend, such as the macroeconomic environment, regional supply–demand structure, and city-level premiums, are not adequately captured by the current representation. Thus, Vd15 both demonstrates the value of textual semantics and exposes the capability boundary of the existing framework, which is precisely what the CAR framework in Section 5 aims to address.

## 5. CAR

### 5.1. Architecture Design

The five-axis ablation in Section 4 ultimately identifies a unified backbone: the independent-projection fusion architecture (C2), the reconstruction objective (G1), dimension-wise normalization (per_dim_norm), 256-dimensional embeddings, and full-dimensional, low-weight text alignment (TE2, InfoNCE, w=0.1). However, cross-domain generalization remains only partially resolved. Under cross-tier extrapolation (E3), several socioeconomic tasks, such as spatial morphology and nighttime-light statistical labels, still yield negative $R^2$ values. Further analysis shows that the problem does not originate from the input side. C2 already provides an independent representation space for each data stream through per-stream projection. However, during training, a single shared decoder is still used to reconstruct all 119 input dimensions. As a result, high-dimensional and high-variance data streams dominate the optimization process, while some low-dimensional semantic information is insufficiently preserved, eventually leading to structural degradation under cross-domain settings.

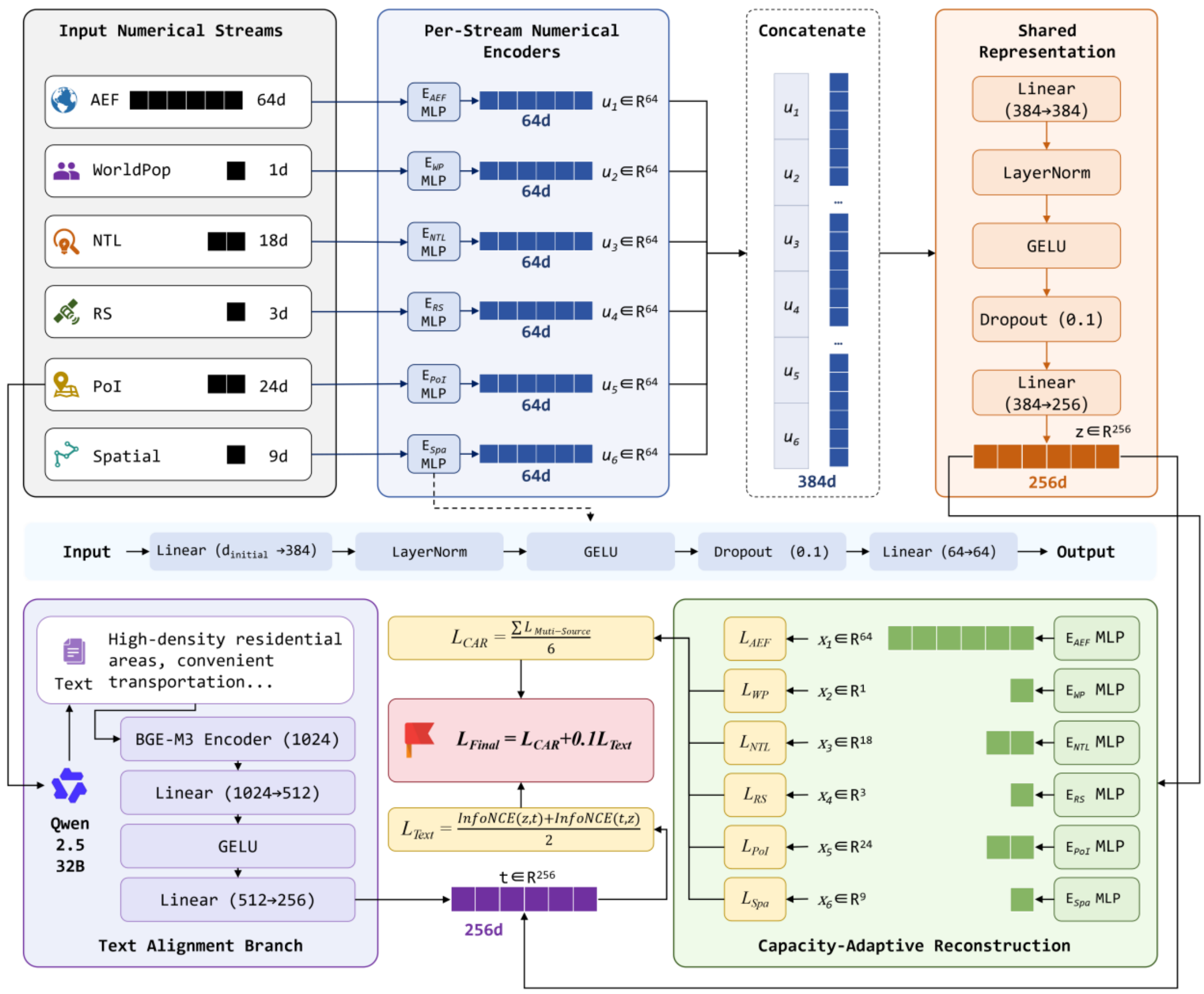

**Fig. 5. Architecture of the Capacity-Adaptive Reconstruction (CAR) framework.**

To address this issue, we propose the CAR framework. CAR does not redesign the entire encoder; instead, it builds on the optimal backbone identified in Section 4 to resolve the problem that capacity fairness exists only on the input side. Specifically, CAR retains the per-stream encoders, shared representation structure, and TE2 text-alignment mechanism of C2, and modifies only the reconstruction stage. The original shared decoder is replaced with per-stream independent decoders, and the unified reconstruction loss is replaced with stream-level reconstruction losses. In this way, each data stream is assigned independent capacity not only before entering the shared representation space, but also during reconstruction through its own optimization objective. CAR therefore extends capacity fairness from the input side to the entire reconstruction process. The overall architecture is shown in Fig. 5.

#### 5.1.1. Input Side: Numerical-Stream Encoders

The input-side design follows the C2 architecture identified in Section 4. Each sample consists of six numerical streams: AEF, WorldPop, NTL, RS, POI, and Spatial, with original dimensionalities of 64, 1, 18, 3, 24, and 9, respectively. To prevent different streams from becoming imbalanced before fusion because of their original dimensional differences, CAR assigns an independent encoder, $\mathrm{E_s}$, to each data stream and projects all streams into a unified 64-dimensional subspace:

$$\mathrm{u_s} = \mathrm{E_s}(\mathrm{x_s}), \quad \mathrm{u_s} \in \mathbb{R}^{64}, \quad \mathrm{s} = 1, \dots, 6 \tag{9}$$

Each $\mathrm{E_s}$ is implemented as a two-layer MLP:

$$\mathrm{Linear}(\mathrm{d_s} \rightarrow 64) \rightarrow \mathrm{LayerNorm} \rightarrow \mathrm{GELU} \rightarrow \mathrm{Dropout}(0.1) \rightarrow \mathrm{Linear}(64 \rightarrow 64)$$

This design ensures capacity fairness on the input side: regardless of differences in original dimensionality, all data streams occupy the same 64-dimensional bandwidth before entering the shared representation space.

#### 5.1.2. Shared Representation and Text Alignment

The six 64-dimensional stream representations are first concatenated into a 384-dimensional vector and then compressed into the final embedding through a shared trunk:

$$\mathrm{z} = \mathrm{T}([\mathrm{u_1}; \mathrm{u_2}; \cdots; \mathrm{u_6}]), \quad \mathrm{z} \in \mathbb{R}^{256} \tag{10}$$

The shared trunk is structured as follows:

$$\mathrm{Linear}(384 \rightarrow 384) \rightarrow \mathrm{LayerNorm} \rightarrow \mathrm{GELU} \rightarrow \mathrm{Dropout}(0.1) \rightarrow \mathrm{Linear}(384 \rightarrow 256)$$

The resulting 256-dimensional vector, z, is the final representation output by CAR. The text stream does not participate in the forward computation of the trunk; instead, it is mapped to the same 256-dimensional space as z through an independent projection head:

$$\mathrm{t} = \mathrm{P_{text}}(\mathrm{e_{text}}), \quad \mathrm{t} \in \mathbb{R}^{256} \tag{11}$$

where $P_{\text{text}}$ is a two-layer MLP:

$$\mathit{Linear}(1024 \rightarrow 512) \rightarrow \mathit{GELU} \rightarrow \mathit{Linear}(512 \rightarrow 256) \tag{12}$$

$\mathrm{e_{text}}$ denotes the BGE-M3 embedding of the LLM-abstracted functional-zone description (Section 3.2.1). A bidirectional InfoNCE loss is used between the textual and numerical embeddings:

$$\mathcal{L}_{\text{text}} = \frac{1}{2}[\mathrm{InfoNCE}(\mathrm{z} \rightarrow \mathrm{t}) + \mathrm{InfoNCE}(\mathrm{t} \rightarrow \mathrm{z})] \tag{13}$$

Text is used only as a semantic constraint during training, and no text input is required during inference.

#### 5.1.3. Output Side: Capacity-Adaptive Reconstruction

The key difference between CAR and the backbone in Section 4 lies on the output side. In the original G1 framework, all 119 input dimensions are reconstructed jointly by a single shared decoder. Although capacity balancing has already been achieved on the input side, different data streams still compete within the same reconstruction objective, allowing high-dimensional and high-variance streams to more easily dominate the optimization process. CAR eliminates this competition by assigning an independent decoder to each data stream:

$$\hat{x}_s = D_s(z), \quad \hat{x}_s \in \mathbb{R}^{d_s} \tag{14}$$

Each $D_s$ is implemented as a two-layer MLP:

$$\mathrm{Linear}(256 \rightarrow 256) \rightarrow \mathrm{GELU} \rightarrow \mathrm{Linear}(256 \rightarrow \mathrm{d_s}) \tag{15}$$

In this way, the six data streams are reconstructed through six independent output branches, with each stream assigned its own optimization objective and error-feedback pathway. The reconstruction loss for stream s is defined as the dimension-wise normalized MSE:

$$\mathcal{L}_s = \frac{1}{d_s}\sum_{k=1}^{d_s}\left(\frac{\hat{x}_{s,k} - x_{s,k}}{\sigma_{s,k}}\right)^2 \tag{16}$$

where $\sigma_{s,k}$ denotes the standard deviation of the corresponding dimension within the current batch, which is used to remove scale differences and implement per_dim_norm.

The final numerical reconstruction loss is defined as the simple average of the losses over the six streams:

$$\mathcal{L}_{\mathrm{CAR}} = \frac{1}{6}\sum_{s=1}^{6}\mathcal{L}_s \tag{17}$$

Compared with shared reconstruction, this design ensures that each data stream receives equal weight in the loss function, thereby extending capacity fairness from the encoding stage to the reconstruction stage.

#### 5.1.4. Total Training Loss

The total loss of CAR consists of the numerical reconstruction loss and the text-alignment loss:

$$\mathcal{L} = \mathcal{L}_{\mathrm{CAR}} + w \cdot \mathcal{L}_{\mathrm{text}}, \quad w = 0.1 \tag{18}$$

The numerical reconstruction loss preserves the informational completeness of the multi-source data, while the text-alignment loss provides semantic constraints that are transferable across cities. Joint optimization of these two losses shapes the final 256-dimensional representation space.

### 5.2. Validation Experiment Design

To verify whether the performance gains of CAR indeed arise from the per-stream decoders and stream-level reconstruction mechanism, rather than from increased parameter size or additional training signals, this section designs five controlled experiments in three categories. First, we use the optimal configuration identified in Section 4 as a strong baseline: the C2 encoder, G1 reconstruction objective, TE2 text alignment with full-dimensional InfoNCE and w=0.1, per_dim_norm, and 256-dimensional embeddings. Its output side still uses a shared decoder. Second, we construct the main CAR model. With the input side, text constraint, and embedding dimensionality kept unchanged, CAR only replaces the shared decoder with per-stream decoders and replaces the unified reconstruction loss with stream-level reconstruction losses, thereby directly testing the effectiveness of the core CAR design. Finally, we construct two types of extended variants. One adds a cross-stream contrastive loss to CAR with $\lambda$=0.1 and 0.5, to test whether additional training signals provide further gains. The other deepens the decoder by expanding the two-layer MLP to a three-layer MLP, to test whether the performance improvement is merely due to increased parameter size. In this way, the structural contribution, training-signal contribution, and parameter-size contribution of CAR can be clearly distinguished.

### 5.3. Experimental Results

The overall results of CAR and its variants are reported in Table 24. Compared with the previous best configuration, the main CAR model improves performance under all three split protocols. E1 $R^2$ increases from 0.896 to 0.923, indicating that introducing output-side capacity constraints does not compromise in-distribution fitting ability. E2 $R^2$ increases from 0.832 to 0.848, and ρ increases from 0.894 to 0.901, demonstrating that CAR improves cross-regional transferability. E3 $R^2$ increases from 0.671 to 0.693, and $\rho$ increases from 0.894 to 0.902, indicating that the proposed structure maintains stable gains even under the most stringent cross-tier extrapolation setting.

**Table 24. Overall performance of CAR and its variants under three split protocols.**

| Strategy | E1 | | E2 | | E3 | |
|---|---|---|---|---|---|---|
| | $R^2$ | $\rho$ | $R^2$ | $\rho$ | $R^2$ | $\rho$ |
| Previous best configuration | 0.896 | 0.942 | 0.832 | 0.894 | 0.671 | 0.894 |
| CAR | **0.923** | **0.951** | **0.848** | **0.901** | **0.693** | **0.902** |
| CAR+S2,λ=0.5 | 0.891 | 0.928 | 0.842 | 0.895 | 0.657 | 0.878 |
| CAR+S2,λ=0.1 | 0.892 | 0.930 | 0.846 | 0.899 | 0.670 | 0.881 |
| CAR+ deeper decoder | 0.890 | 0.927 | 0.835 | 0.892 | 0.645 | 0.875 |

The classification and retrieval results also support this conclusion. In the Vd4 classification task (Table 25), CAR achieves higher κ values than the previous best configuration under all three split protocols, increasing from 0.861 to 0.868 under E1, from 0.710 to 0.742 under E2, and from 0.704 to 0.712 under E3. F1 improves slightly under E1 and E3, with E3 increasing from 0.521 to 0.526, although E2 F1 decreases marginally from 0.514 to 0.507. The consistent improvement in κ indicates that CAR embeddings enhance inter-class separability in the classification task, while the slight decrease in E2 F1 may result from boundary fluctuations in certain classes under the cross-regional setting. In the Vd16 retrieval task (Table 26), CAR improves Recall@8 from 0.399 to 0.408 and MRR from 0.642 to 0.650. These results show that the benefits of CAR are not limited to regression labels: both the retrieval task (Vd16) and the κ metric in the classification task improve consistently, with only a slight decrease in Vd4 E2 F1.

**Table 25. Vd4 performance of CAR and variants.**

| Strategy | E1 F1 | E1κ | E2 F1 | E2κ | E3 F1 | E3κ |
|---|---|---|---|---|---|---|
| Previous best configuration | 0.642 | 0.861 | 0.514 | 0.710 | 0.521 | 0.704 |
| CAR | **0.648** | **0.868** | **0.507** | **0.742** | **0.526** | **0.712** |
| CAR+S2,λ=0.5 | 0.617 | 0.798 | 0.438 | 0.686 | 0.484 | 0.646 |
| CAR+S2,λ=0.1 | 0.616 | 0.798 | 0.436 | 0.668 | 0.462 | 0.605 |
| CAR+ deeper decoder | 0.619 | 0.797 | 0.445 | 0.693 | 0.485 | 0.647 |

The extended experiments further show that the performance gain of CAR mainly comes from its structural design. After adding the cross-stream contrastive loss, the overall performance is lower than that of the main CAR model under both high- and low-weight settings: E3 $R^2$ drops to 0.657 in the high-weight version and to 0.670 in the low-weight version, both below the 0.693 achieved by CAR. Vd4 and Vd16 also show corresponding declines. Deepening the decoder likewise brings no additional benefit, with E3 $R^2$ reaching only 0.645 and Vd16 MRR decreasing to 0.584. These results indicate that, within the scope of the current experiments, no extended variant consistently outperforms CAR. The main benefit of CAR does not arise from a larger parameter scale or additional training signals, but from the structural design of per-stream decoders and stream-level reconstruction losses.

**Table 26. Vd16 performance of CAR and variants.**

| Strategy | Recall@8 | MRR |
|---|---|---|
| Previous best configuration | 0.399 | 0.642 |
| CAR | **0.408** | **0.650** |
| CAR+S2,λ=0.5 | 0.393 | 0.591 |
| CAR+S2,λ=0.1 | 0.381 | 0.574 |
| CAR+ deeper decoder | 0.389 | 0.584 |

The label-level results in Table 27 further reveal the mechanism through which CAR operates. The most pronounced improvements occur for labels associated with low-dimensional data streams. For population density, Vd8, which corresponds to the one-dimensional WorldPop input, E3 $R^2$ increases from 0.956 to 0.983. For spatial morphology, Vd13, which corresponds to the nine-dimensional Spatial input, E3 $R^2$ increases sharply from $-1.374$ to 0.806. Vd13 is particularly informative: under the shared decoder, it exhibits clear cross-domain collapse, whereas CAR restores it to a normal performance range through independent reconstruction outputs. This directly verifies the role of output-side capacity fairness in preserving information from low-dimensional semantic streams. A similar recovery from collapse is observed for Vd10, the 18-dimensional nighttime-light stream, where E3 $R^2$ increases from $-0.707$ to 0.135. Labels with already strong baselines also show stable gains; for example, Vd11, POI density, improves from 0.868 to 0.903 under E3. The most notable case is Vd15, housing price: E2 $R^2$ increases from 0.278 to 0.483, and E3 $R^2$ increases from 0.007 to 0.317. Section 4 has shown that text alignment can recover this task from a catastrophic E3 collapse of $-14.439$ to values near zero, but it remains around zero and fails to produce effective prediction. With TE2 text alignment kept unchanged, CAR enables Vd15 to clearly exceed zero for the first time, indicating that this gain comes from improved representation organization rather than from the introduction of additional semantic information.

**Table 27. Per-label $R^2$ comparison between the five-axis baseline and CAR on representative labels, highlighting collapse recovery.**

| Vd | Label | Baseline E1 | CAR E1 | Baseline E2 | CAR E2 | Baseline E3 | CAR E3 |
|---|---|---|---|---|---|---|---|
| Vd8 | Population density | 0.966 | **0.985** | 0.955 | **0.985** | 0.956 | **0.983** |
| Vd10 | Nighttime lights 18d | 0.969 | **0.973** | 0.947 | **0.956** | −0.707 | **0.135** |
| Vd11 | POI density | 0.917 | **0.940** | 0.890 | **0.930** | 0.868 | **0.903** |
| Vd12 | POI distance | 0.992 | **0.999** | 0.992 | **0.999** | 0.985 | **0.999** |
| Vd13 | Spatial morphology | 0.944 | **0.977** | 0.866 | **0.914** | −1.374 | **0.806** |
| Vd15 | Housing price | 0.876 | **0.973** | 0.278 | **0.483** | 0.007 | **0.317** |

## 6. AEF-Econ

### 6.1. Product Description

Based on the trained CAR encoder, we construct the AEF-Econ urban embedding product. The product covers 14.4 million 100-m pixel samples across 36 cities from 2017 to 2024. Each sample is jointly encoded from six numerical data streams: AEF, WorldPop, NTL, RS, POI, and Spatial. CAR natively outputs 256-dimensional embeddings, and we additionally provide 128-dimensional and 64-dimensional compressed versions based on principal component analysis (PCA), with cumulative explained variance ratios of 99.86% and 97.20%, respectively. Because the encoder parameters are fixed during inference, the product can be directly extended to larger spatial extents. In this study, we first generate a 14.4-million-sample version following the same sampling protocol as in Sections 4 and 5, ensuring that the product specification is strictly aligned with the experimental results reported above. The product specifications are summarized in Table 28.

**Table 28. AEF-Econ product specifications.**

| Item | Specification |
|---|---|
| Spatiotemporal coverage | 36 cities, 2017–2024 |
| Sample size | 14.4 million |
| Resolution | 100 m |
| Embedding dimensionality | 256 d(main version) |
| Compressed versions | 128d (99.86%) and 64d (97.20%) |
| Storage size | 13.7 GB / 6.9 GB / 3.4 GB |
| Encoder | CAR_v2 |

### 6.2. Product Self-Diagnostic Assessment

After validating downstream performance, it is also necessary to examine whether the embeddings themselves are well formed. This section evaluates the product from three perspectives: dimensional health, city-tier consistency, and intra-urban spatial structure.

### 6.2.1. Embedding Dimensional Health

To assess the embedding quality of AEF-Econ, we diagnose it from three aspects: dimensional collapse, distributional shape, and cross-city discriminative power. The results are shown in Fig. 6.

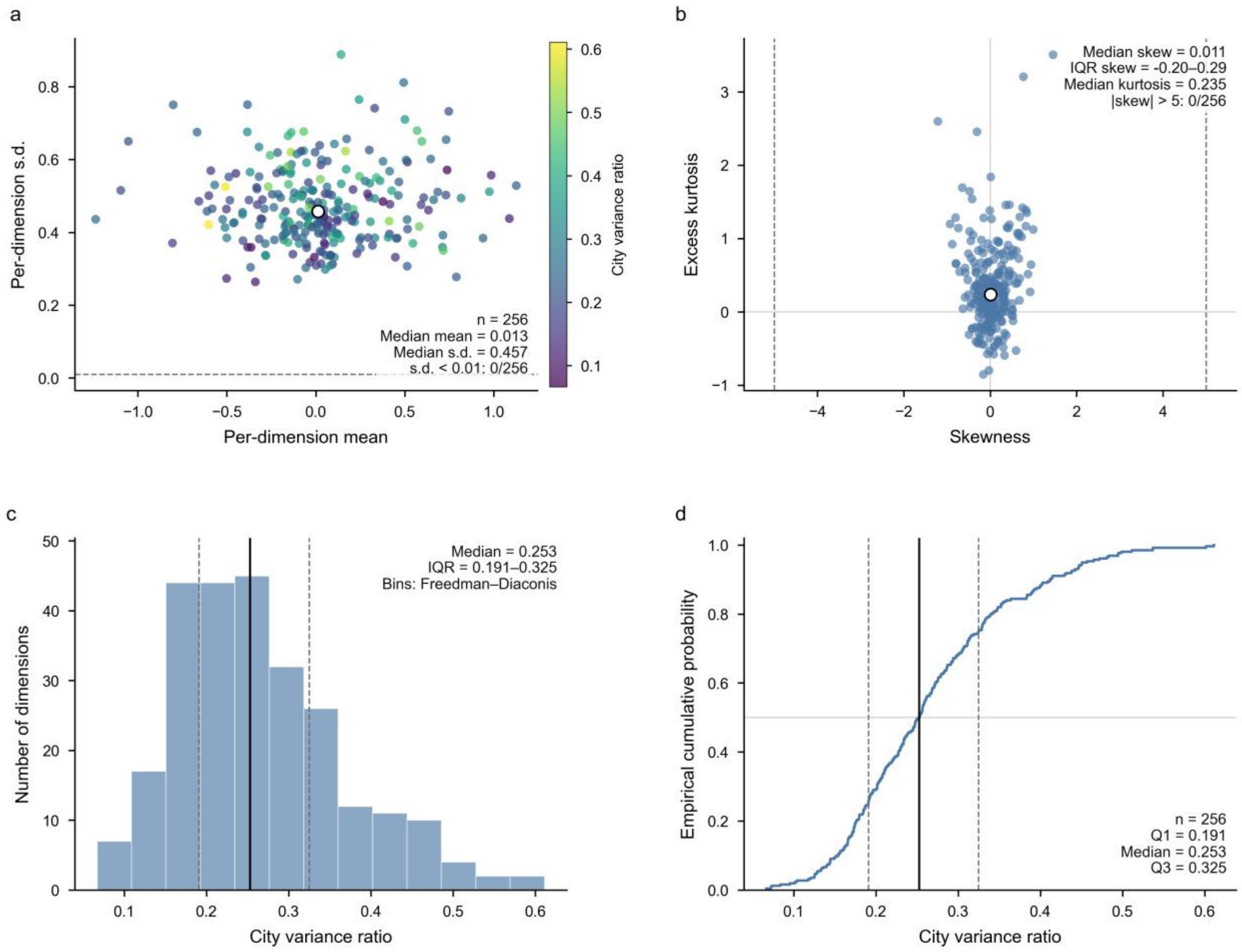

**Fig. 6. Embedding dimension health diagnostics of AEF-Econ (256 dimensions).** (a) Per-dimension mean vs. standard deviation, colored by cross-city variance ratio; no dimension falls below the collapse threshold (s.d. < 0.01). (b) Skewness vs. excess kurtosis; all dimensions remain within normal bounds. (c) Histogram of cross-city variance ratio. (d) Empirical cumulative distribution of cross-city variance ratio.

We first examine dimensional collapse. If the standard deviation of a dimension across all samples is close to zero, that dimension cannot distinguish among samples and thus represents a typical case of dimension collapse. The literature commonly uses std < 0.01 as the diagnostic threshold. For the 256 dimensions of AEF-Econ, the median standard deviation is 0.46 and the minimum is 0.18, both far above this threshold (Fig. 6a). This indicates that no collapsed dimensions are present and that all dimensions contribute to information representation.

We next examine the distributional shape. Skewness and excess kurtosis are used to assess whether the embeddings are dominated by a small number of anomalous samples. For AEF-Econ, the median skewness is 0.011 and the median excess kurtosis is 0.235, while the skewness values of all dimensions fall within the range of [−1.5, 1.5] (Fig. 6b), far from the commonly used abnormality threshold of |skew|=5. This suggests that the embedding distribution is generally stable and contains no obvious pathological dimensions.

Finally, we evaluate cross-city discriminative power. The cross-city variance ratio is defined as follows:

$$R = \frac{\sigma^2_{between-city}}{\sigma^2_{between-city} + \sigma^2_{within-city}} \tag{19}$$

Here, R→1 indicates that the embeddings predominantly encode city identity, whereas R→0

suggests excessive mixing of different cities in the embedding space. For AEF-Econ, the median proportion of cross-city variance was 0.25, with an interquartile range of [0.19, 0.32], and values across all dimensions ranged from 0.07 to 0.62 (Fig. 6c–d). These results indicate that most embedding dimensions simultaneously preserve inter-city differences and intra-city heterogeneity, thereby achieving a balance between cross-city separability and within-city resolution.

### 6.2.2. Consistency with the Urban Hierarchy

To examine whether the embedding space spontaneously captured the hierarchical structure of cities, we averaged the 256-dimensional embeddings of 50,000 pixels for each city to obtain city-level embedding vectors. Inter-city similarity was then quantified using cosine similarity. Subsequently, within-group and between-group similarities were calculated according to the urban hierarchy defined in Section 3.1, comprising T1, T1.5, T2, T3, and T4 cities.

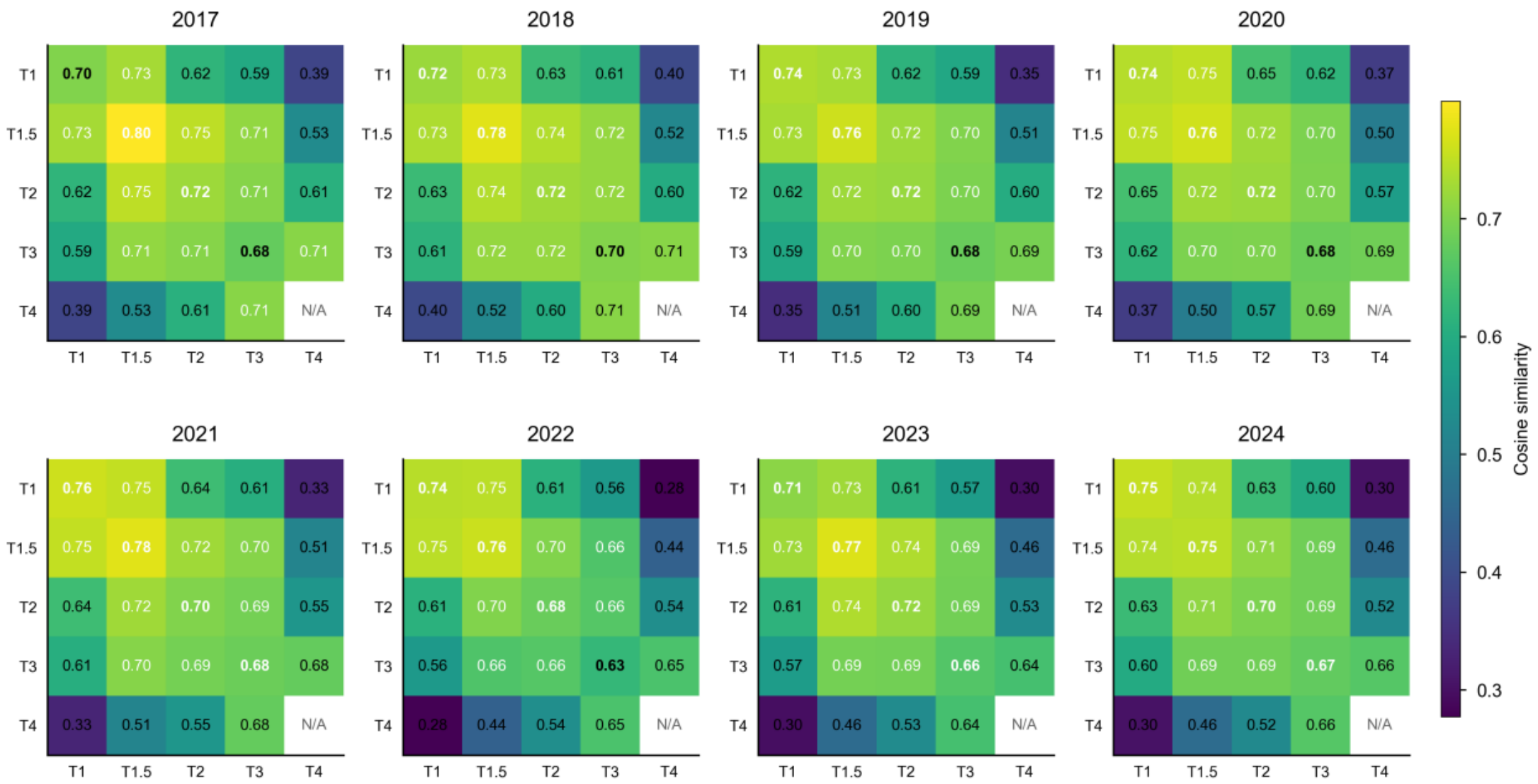

**Fig. 7. Tier-level cosine similarity matrices of city-mean embeddings from 2017 to 2024, showing self-organized hierarchical structure (T1 > T1.5 > T2 > T3 > T4) without any tier label supervision.**

Fig. 7 presents the urban-tier similarity matrices for eight consecutive years from 2017 to 2024. The results reveal a highly consistent hierarchical structure: cities within the same tier consistently exhibit the highest similarity, while within-tier similarity decreases progressively from higher to lower tiers. In contrast, cross-tier similarity declines steadily as the tier gap widens. Taking 2024 as an example, the mean within-tier similarity follows the order:

$$s_{T1} = 0.7545 > s_{T1.5} = 0.7460 > s_{T2} = 0.6973 > s_{T3} = 0.6742 \quad (20)$$

In contrast, cross-tier similarity gradually decreased from 0.74 for T1↔T1.5 to 0.30 for T1↔T4. This pattern remained stable across all eight years, indicating that the observed hierarchical structure was not a random artifact but rather a persistent organizational property of the embedding space. Notably, no urban-tier labels were used during CAR training. Therefore, this stable tiered stratification can only be attributed to the model's self-organization of multi-source urban signals. These results demonstrate that AEF-Econ not only captures intra-city heterogeneity but also learns, at the city scale, integrated socioeconomic characteristics that are consistent with the real-world urban hierarchy.

### 6.2.3. Stability of Intra-city Spatial Structure

To examine whether the embeddings preserve intra-city spatial structure, we selected six representative cities—Beijing (T1), Shenzhen (T1), Wuhan (T1.5), Kunming (T2), Xi'an (T2), and Lhasa (T4). For each city, 5,000 pixels were randomly sampled from the 2024 dataset, and their 256-dimensional embeddings were projected into a two-dimensional space using Uniform Manifold Approximation and Projection (UMAP) (Mcinnes et al., 2018). Because this analysis focuses on whether the embeddings form a continuous and organized low-dimensional manifold, rather than on explaining total variance, we adopted this nonlinear dimensionality-reduction method as it is more suitable for embedding diagnostics. UMAP was fitted separately for each city, and the results were used solely to analyze the internal structure of individual cities.

However, the manifold geometry alone is insufficient to determine whether the learned structure is substantively meaningful. Therefore, NTL intensity was further introduced as an external reference. NTL is widely used as a proxy for economic activity (Elvidge et al., 2017). If the embeddings indeed capture intra-city socioeconomic differences, regions with high and low NTL intensity should exhibit a continuous distribution on the manifold rather than being randomly intermingled.

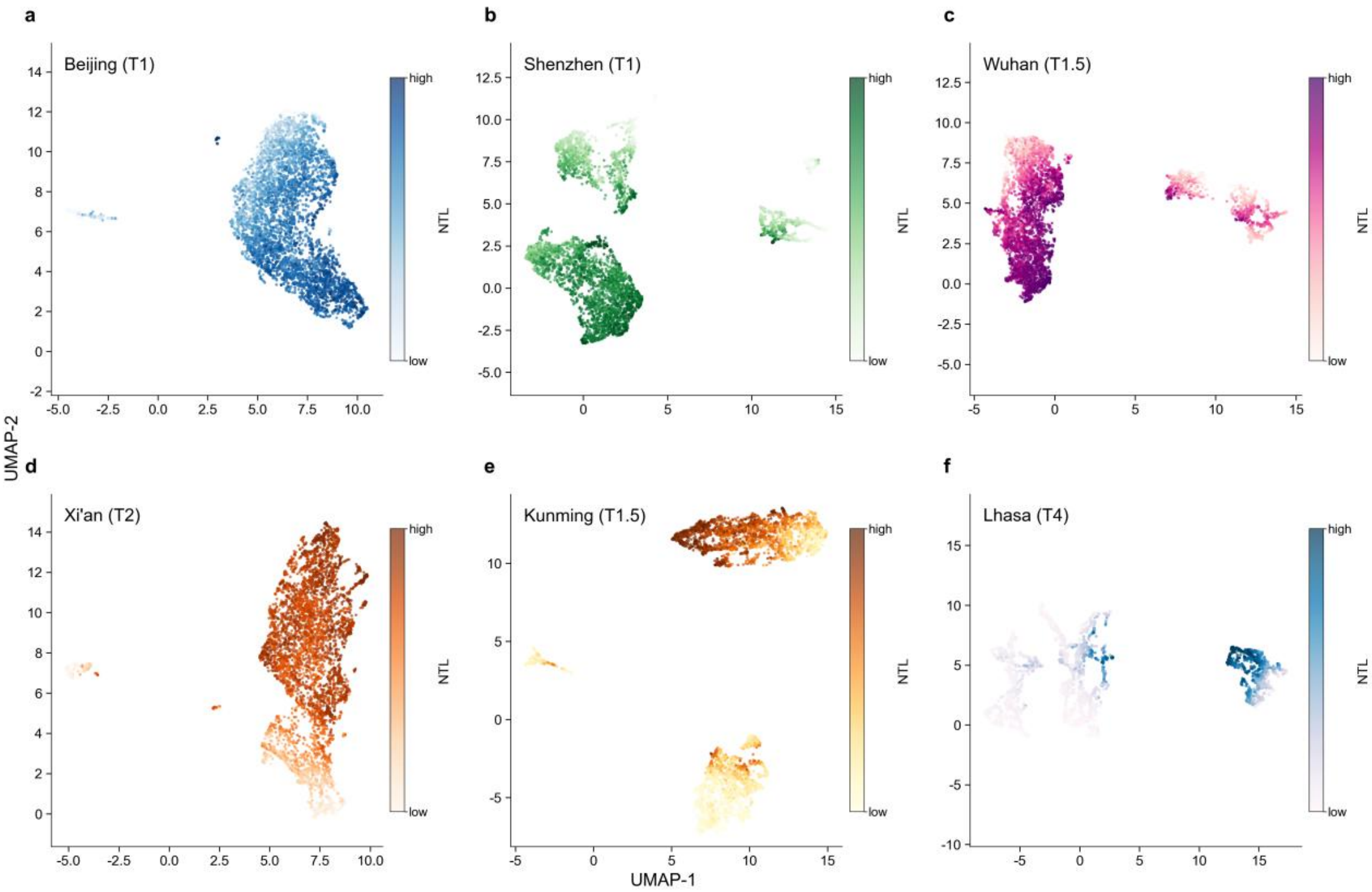

**Fig. 8. UMAP projections of intra-city pixel embeddings for six representative cities (Beijing, Shenzhen, Wuhan, Kunming, Xi'an, Lhasa), colored by nighttime light intensity, showing continuous NTL-aligned manifold structures.**

The results (Fig. 8) show that all six cities formed clear and continuous low-dimensional manifolds, rather than random scatter patterns or collapsed single-point structures. The manifold geometry varied markedly across cities: Beijing and Xi'an exhibited relatively continuous dominant manifolds, whereas Wuhan, Shenzhen, Kunming, and Lhasa showed multiple subclusters or connected components, reflecting distinct modes of spatial organization. When colored by NTL intensity, each city further displayed an economic gradient consistent with its manifold structure, with areas of high and low economic activity transitioning continuously along the manifold rather than being randomly distributed. For example, Beijing and Xi'an exhibited pronounced core–periphery gradients, while the multiple subclusters observed in Wuhan and Kunming corresponded to economically active areas at different hierarchical levels.

## 6.3. Application Cases

### 6.3.1. Case 1: Urban Growth Dynamics

AEF-Econ covers eight consecutive years from 2017 to 2024, enabling urban growth processes to be characterized directly from the temporal evolution of the embeddings. For a pixel location (p), its eight-year embedding sequence is denoted as $(z_p^{2017}, z_p^{2018}, \ldots, z_p^{2024})$, and the embedding trajectory length is defined as: $L_p = \sum_{t=2017}^{2023} \left| z_p^{t+1} - z_p^{t} \right|_2$. This metric measures the cumulative displacement of a pixel in the socioeconomic embedding space. A larger trajectory length indicates more pronounced joint changes in multi-source features, including the built environment, NTL, POIs, and spatial structure, during the study period; conversely, a smaller trajectory length suggests that the area remained relatively stable. Unlike conventional growth indicators, the trajectory length is defined entirely from the embeddings themselves and does not require growth labels or manually specified growth criteria.

When computed for 1.8 million pixels across 36 cities, the trajectory length exhibited a distinctly right-skewed, long-tailed distribution, with a median of 61.2, a 99th percentile of 90.3, and a maximum of 144.5. This pattern suggests that urban growth was concentrated primarily in a limited number of rapidly reorganizing areas, rather than occurring uniformly across the entire urban space.

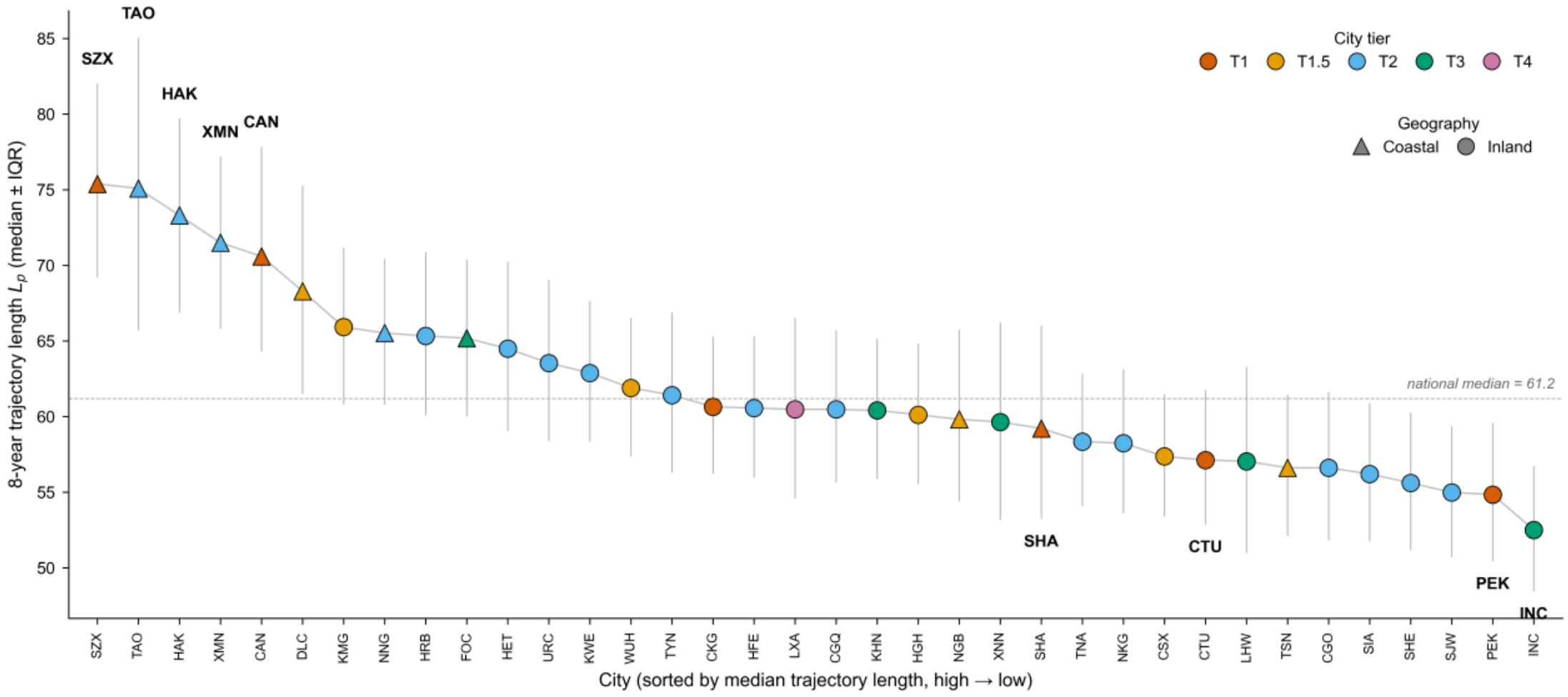

**Fig. 9. Ranking of 36 cities by median embedding trajectory length (2017–2024), revealing that coastal cities dominate the top positions while city tier does not simply predict growth intensity.**

Fig. 9 further presents the ranking of growth intensity across the 36 cities. A notable pattern is that the top five cities—Shenzhen, Qingdao, Haikou, Xiamen, and Guangzhou—are all coastal or island cities, with median trajectory lengths exceeding 70, substantially higher than the national median of 61.2. This finding is highly consistent with the sustained development of major coastal development zones during the study period, including the Qianhai Cooperation Zone, the West Coast New Area, the Jiangdong New Area, the Second Eastern Coastal Area, and the Nansha New Area. In contrast, urban tier and growth intensity do not exhibit a simple one-to-one correspondence: Shenzhen ranks first, whereas Shanghai, Chengdu, and Beijing appear in the lower part of the ranking. This suggests that AEF-Econ captures actual processes of spatial reorganization rather than city size or administrative hierarchy per se.

Table 29 provides further external validation. In areas containing well-known new districts, such as the Qianhai Cooperation Zone, Lingang Special Area, Tianfu New Area, Future Sci-Tech City, Tongzhou Sub-Center, and Wuhan Optics Valley, the mean values of the top five high-trajectory-length pixels all exceed 87, markedly higher than the national mean of 62.18. Among them, the Qianhai Cooperation Zone (119.88), Lingang Special Area (109.90), and Tianfu New Area (106.28) rank at the top. Because the development trajectories of these new districts were not used during model training, this result indicates that AEF-Econ can spontaneously identify real-world priority development areas using embedding trajectories alone, thereby revealing the spatial patterns of urban growth and functional reorganization.

**Table 29. Top-5 mean trajectory length at six known new development zones.**

| City | New District / New Town | Coordinates (lon, lat) | Mean Trajectory Length of the Top Five Pixels |
|---|---|---|---|
| SZX Shenzhen | Qianhai Cooperation Zone | (113.91,22.55) | **119.88** |
| SHA Shanghai | Lingang Special Area | (121.91,30.86) | **109.90** |
| CTU Chengdu | Tianfu New Area | (104.07,30.43) | **106.28** |
| HGH Hangzhou | Future Sci-Tech City | (120.04,30.32) | 92.77 |
| PEK Beijing | Tongzhou Sub-Center | (116.66,39.91) | 89.39 |
| WUH Wuhan | Optics Valley | (114.43,30.46) | 87.16 |

### 6.3.2. Case 2: Unsupervised Cross-city Functional-zone Clustering and Multi-dimensional Diagnostics

AEF-Econ not only represents individual pixels but also supports the discovery of spatial organization at the national scale. To this end, we applied MiniBatch K-means clustering to the 256-dimensional embeddings of all 1.8 million pixels in 2024. The main analysis used k=16, with k=8 and k=24 adopted for robustness checks. During clustering, only the embedding vectors themselves were used, without incorporating any labels related to functional zones, urban tiers, or regions. Five types of functional anchors—central business districts (CBDs), old urban cores, railway stations, universities, and new districts—were used solely for post hoc interpretation.

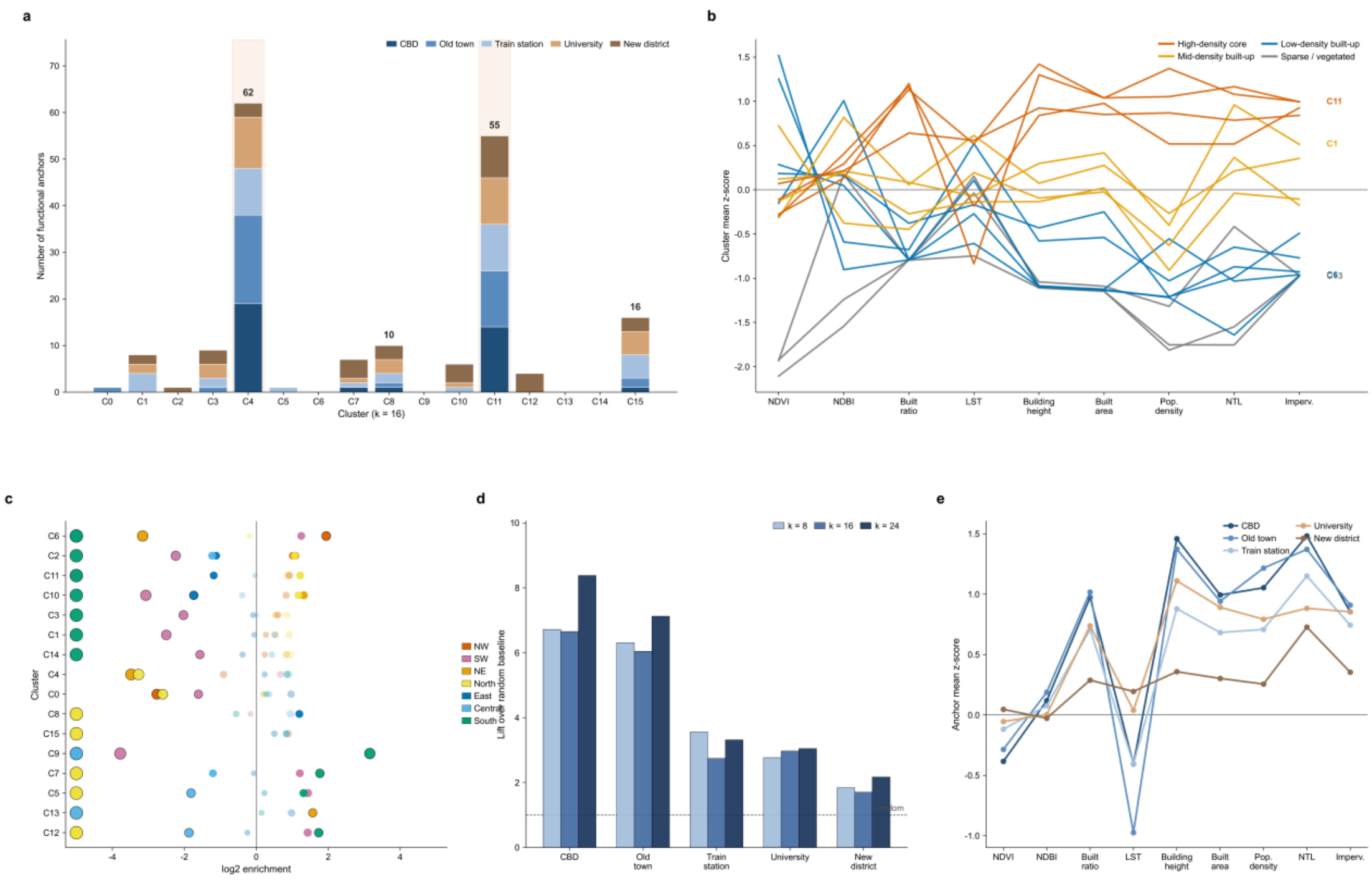

**Fig. 10. Cross-city unsupervised clustering (K-means, k=16) of 1.8 million pixel embeddings (2024).** (a) Functional anchor distribution across clusters. (b) Cluster-mean z-scores on nine physical and socioeconomic attributes. (c) Regional enrichment per cluster. (d) Anchor-to-cluster concentration ratio at k=8, 16, 24. (e) Mean attribute profiles of five anchor types.

Fig. 10 demonstrates the real-world interpretability of the clustering results from five complementary perspectives. First, the functional anchors were not randomly dispersed but were instead highly concentrated in a small number of clusters: C4 and C11 contained 62 and 55 anchors, respectively, together accounting for more than 65% of all anchors. CBDs and old urban cores showed the strongest concentration in these clusters (Fig. 10a), indicating that the embeddings spontaneously identified urban core areas. In terms of attribute profiles, these clusters did not simply correspond to predefined functional labels. Instead, they formed continuous gradients across NDVI, building height, population density, NTL, and built-up intensity, which can be summarized as high-density urban cores, medium-density built-up areas, low-density built-up areas, and vegetation-dominated areas (Fig. 10b). Meanwhile, clusters with similar built-environment characteristics still preserved regional differences; for example, C4 was more enriched in southern and southwestern China, whereas C11 was more prevalent in northern and northeastern China (Fig. 10c), suggesting that the embeddings also encoded regional context. Cross-city consistency was also strong: CBD and old-core anchors were 6–8 times more likely than random expectation to fall into the same cluster, and this pattern remained stable across the three clustering granularities of k=8, k=16, and k=24 (Fig. 10d). This indicates that the observed structure was not a local phenomenon driven by individual cities. Finally, Fig. 10e explains why CBDs, old urban cores, railway stations, and universities were not further separated: their profiles in building height, population density, NTL, and built-up intensity were highly similar, indicating that they all essentially correspond to high-density built environments. In contrast, new districts were closer to the national average and exhibited greater cross-city variability, resulting in a more dispersed distribution. Overall, the clustering results of AEF-Econ primarily reflect spatial–environmental structure rather than administratively defined functional-zone names.

### 6.3.3. Case 3: Unsupervised Identification of Intra-city Socioeconomic Gradients

AEF-Econ supports not only cross-city comparison but also the characterization of continuous socioeconomic gradients within cities. To demonstrate this capability, we selected four representative cities—Beijing, Shanghai, Wuhan, and Xi'an—and performed intra-city K-means clustering (k=5) separately on their 50,000 pixel-level embeddings from 2024. The clustering procedure used only the 256-dimensional embedding vectors, without incorporating socioeconomic labels, functional-zone information, or spatial-location constraints. We then constructed a composite intensity score, referred to as the core score, using NTL, building height, and housing prices, and reordered the five clusters from L0 to L4. Here, L4 denotes the areas with the highest composite intensity, whereas L0 represents the lowest-level areas.

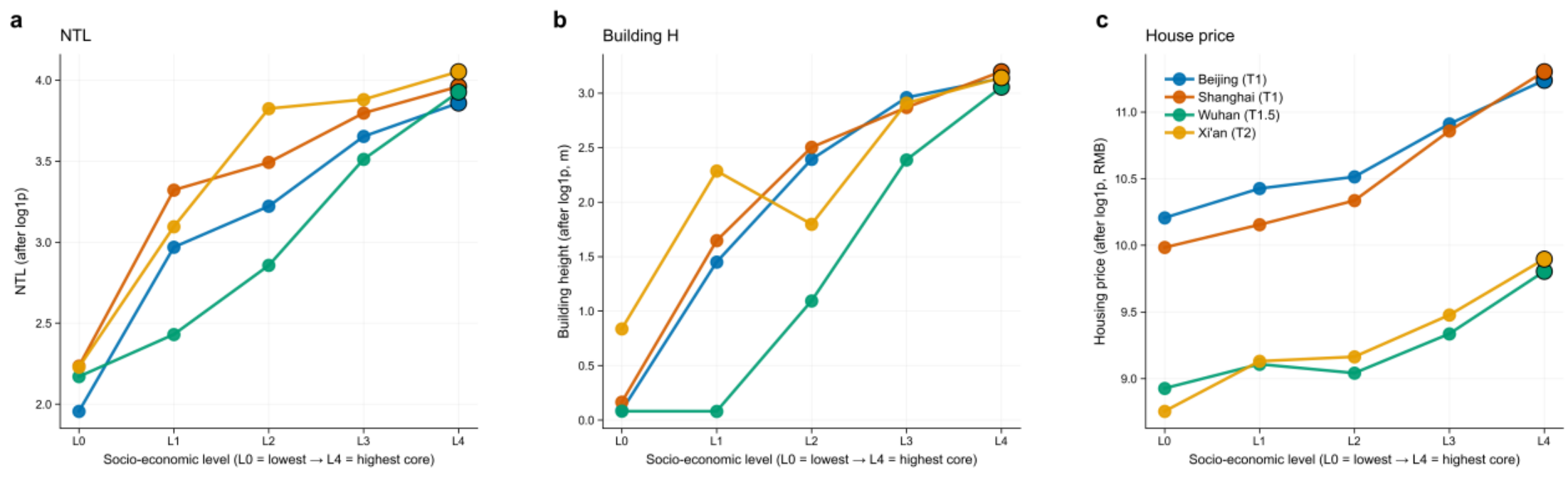

**Fig. 11. Spatial distribution of unsupervised intra-city socioeconomic levels (L0–L4) for Beijing, Shanghai, Wuhan, and Xi'an (2024).** Clustering uses only 256d embeddings; no geographic or land-use information is provided. Rivers and water bodies are naturally recovered as L0 boundaries.

Fig. 11 presents the identified spatial hierarchy. All four cities exhibit clear and continuous spatial gradients rather than random patch-like distributions. Beijing and Xi'an show a typical core–periphery expansion pattern, with higher-level areas concentrated in the central urban districts and gradually transitioning toward the periphery. By contrast, Shanghai and Wuhan display more pronounced polycentric structures, in which the Huangpu River and the Yangtze River are naturally identified as low-intensity boundaries that partition high-intensity areas into multiple clusters. Notably, these spatial structures emerge solely from the embeddings themselves. Although no information on roads, water bodies, or administrative boundaries was used during clustering, the method still recovered realistic intra-city organizational patterns.

Fig. 12 further clarifies the socioeconomic meaning of these hierarchical levels. Across three independent indicators—NTL, building height, and housing prices—all four cities exhibit a consistent gradient, with values generally increasing from L0 to L4. Specifically, NTL reflect economic activity, building height represents the intensity of the built environment, and housing prices capture market value. Although these indicators were not used in the clustering process, they align closely with the hierarchical structure derived from the embeddings. This demonstrates that AEF-Econ recovers not merely an abstract geometric structure, but a real-world socioeconomic gradient within cities.

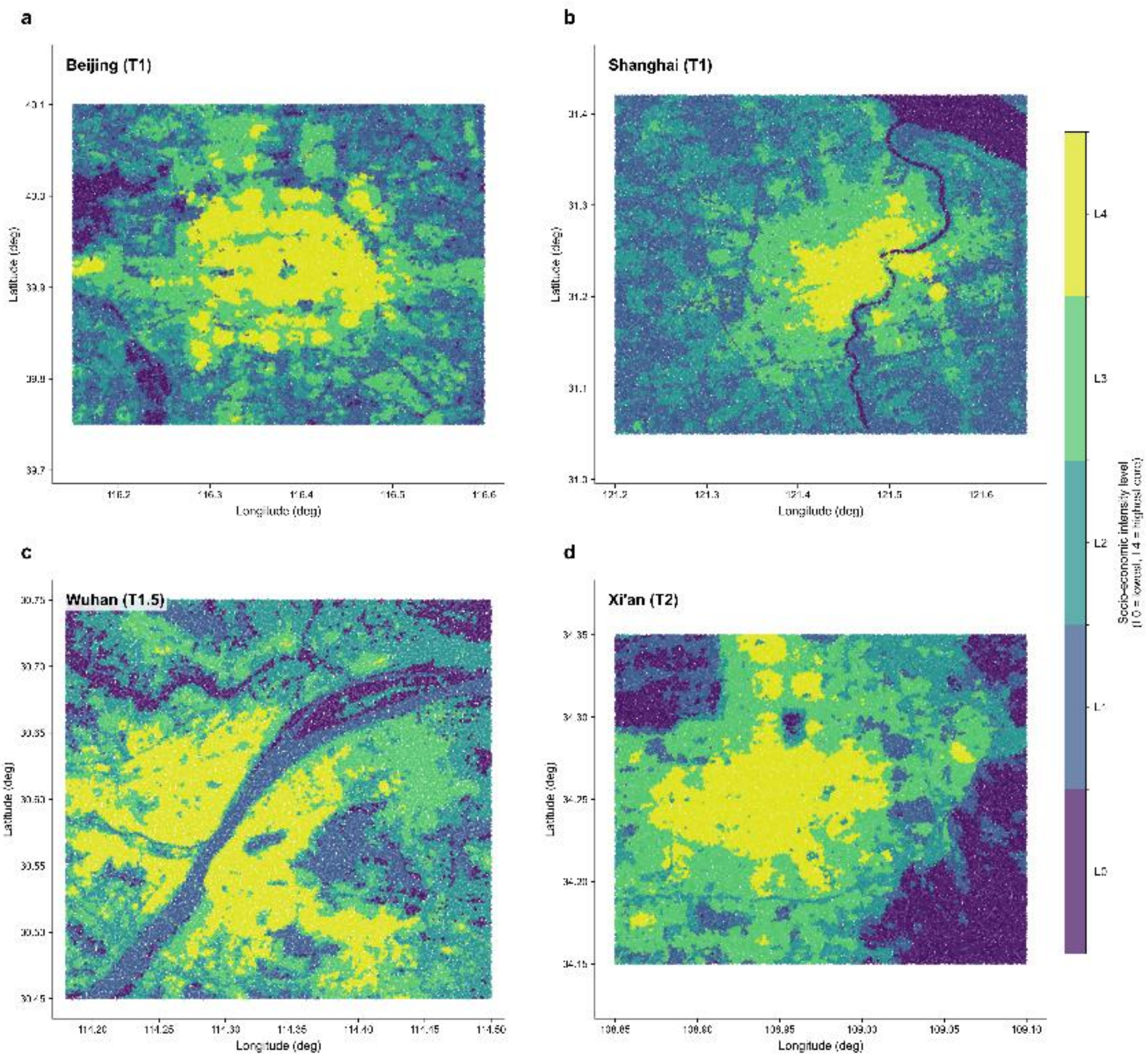

**Fig. 12. Socioeconomic attribute profiles across levels L0–L4 for four cities.** (a) Nighttime light. (b) Building height. (c) Housing price. All three indicators increase monotonically from L0 to L4, confirming that the embedding-derived levels correspond to real socioeconomic gradients.

At the same time, Fig. 12 also preserves inter-city differences. Housing prices in Beijing and Shanghai are generally higher than those in Wuhan and Xi'an across all hierarchical levels, whereas building height and nighttime light intensity remain at comparable levels. This pattern is consistent with real-world observations: core areas in different cities tend to exhibit high built-environment intensity and activity levels, but market value remains influenced by the overall development level of the city. Therefore, the hierarchical structure derived from AEF-Econ can not only characterize intra-city heterogeneity but also preserve the relative scale information required for cross-city comparison.

Another noteworthy pattern is the spatial distribution of the L0 level. As shown in Fig. 11, L0 mainly appears along rivers, water bodies, mountainous areas, and peripheral ecological zones, rather than simply corresponding to distant suburban areas. This suggests that the CAR embeddings primarily identify spatial units with the lowest built-environment intensity and socioeconomic activity, rather than specific land-use categories. Therefore, AEF-Econ has clear socioeconomic semantics; however, further differentiation among specific land-cover classes, such as water bodies, forests, or croplands, would still require dedicated land-cover information.

## 7. Discussion and Limitations

CAR effectively alleviates inter-stream capacity competition through per-stream independent decoding; however, the current framework is subject to three structural constraints. First, reconstructing the six streams with equal weights (1/6) implicitly assumes that all streams contribute equally to the learned representation. In areas where POIs are extremely sparse or nighttime light signals are saturated, this design may force the encoder to fit noise. Moreover, the text stream depends on the coverage of Amap POIs and the LLM-based abstraction procedure, which limits its direct transferability to regions outside China. Second, the bounding boxes of the training samples were manually checked to ensure that they primarily covered core built-up areas and urban–rural transition zones. Consequently, the density of multi-source signals declines substantially in distant agricultural areas, ecological zones, and extremely sparsely populated regions: WorldPop values approach zero, POIs are nearly absent, and NTL signals are close to background noise. The reliability of the embeddings in these areas has not been sufficiently validated, and direct extrapolation should therefore be treated with caution. Finally, Vd15, representing housing prices, achieved an $R^2$ of only 0.317 on E3, indicating that pixel-scale embedding compression cannot substitute for the incorporation of structural variables such as the macroeconomic environment and city-level premiums.

Future work can proceed in three directions. First, adaptive stream-weighting mechanisms should be introduced, and the training domain should be expanded to the national or even global scale. Incorporating samples from non-built-up areas would further improve the applicability of the embeddings across the full spatial domain. Second, temporal encoders or time-contrastive objectives should be explored to replace the current simplified strategy of independently encoding each year, thereby better capturing the path dependence of urban evolution. Third, due to hardware constraints, the present experiments were conducted only at a single spatial resolution of 100 m. The performance of CAR at the native 10 m resolution or at coarser scales, such as 500 m and 1 km, has not yet been evaluated. Multi-scale extension, together with joint training using AEF physical embeddings, represents a key pathway toward constructing a comprehensive dual-channel foundation embedding that integrates physical and socioeconomic representations.

**8. Conclusion**

This study addresses the limited plug-and-play capability of AEF in socioeconomic tasks. Through 31 controlled experiments, we systematically identified the capacity-competition bottleneck in multi-source fusion, where low-dimensional semantic streams are suppressed by high-dimensional streams. To overcome this limitation, we proposed the CAR framework, which extends capacity fairness from the input side to the output side through per-stream independent decoders and stream-level reconstruction losses. The overall framework improved the cross-region $R^2$ from 0.301 with AEF alone to 0.832 with the five-axis base and further to 0.848 with CAR; similarly, the cross-tier $R^2$ increased from 0.160 to 0.671 and then to 0.693. In addition, collapsed labels were restored to a stable range. The CAR-based AEF-Econ product was further validated through three levels of self-diagnostics and three types of urban case studies. Under fully unsupervised conditions, it can spontaneously recover urban hierarchical structures, functional-zone organization, and intra-city socioeconomic gradients, thereby forming a socioeconomic remote-sensing foundation embedding that is dual and complementary to the physical embeddings of AEF.

**Disclosure statement**

No potential conflict of interest was reported by the authors.

**ORCID**

Shuyang Hou: http://orcid.org/0009-0000-6984-9959

Huayi Wu: http://orcid.org/0000-0003-3971-0512

**Acknowledgements**

This work is built entirely upon the open-source contributions of the Google DeepMind and Google Earth Engine teams to AlphaEarth Foundations, for which we express our sincere gratitude. We also gratefully acknowledge the WorldPop team (Bondarenko et al., 2025), the CLCD/GISA team at Wuhan University (Yang and Huang, 2021; Huang et al., 2022), the JRC GHSL team (Corbane et al., 2021), the NOAA VIIRS DNB team (Elvidge et al., 2017), the BAAI BGE team (Chen et al., 2024), the OpenStreetMap community , and all contributors of the public data products that provided input data for this benchmark. The performance limitations of AEF on socioeconomic tasks reported in this study should not be interpreted as evidence that "AEF is ineffective"; rather, they should be understood as demonstrating that any globally trained foundation model requires domain adaptation when applied out of domain. This study presents one concrete pathway for such adaptation.

**Reference**

Bardes, A., Ponce, J., and LeCun, Y.: Vicreg: Variance-invariance-covariance regularization for self-supervised learning, arXiv preprint arXiv:2105.04906, 2021.

Bondarenko, M., Priyatikanto, R., Tejedor Garavito, N., Zhang, W., McKeen, T., Cunningham, A., Woods, T., Hilton, J., Cihan, D., and Nosatiuk, B.: Constrained estimates of 2015-2030 total number of people per grid square at a resolution of 3 arc (approximately 100m at the equator) R2025A version v1, 2025.

Brown, C. F., Kazmierski, M. R., Pasquarella, V. J., Rucklidge, W. J., Samsikova, M., Zhang, C., Shelhamer, E., Lahera, E., Wiles, O., and Ilyushchenko, S.: Alphaearth foundations: An embedding field model for accurate and efficient global mapping from sparse label data, arXiv preprint arXiv:2507.22291, 2025.

Caron, M., Touvron, H., Misra, I., Jégou, H., Mairal, J., Bojanowski, P., and Joulin, A.: Emerging Properties in Self-Supervised Vision Transformers, arXiv preprint arXiv:2104.14294, 2021.

Chen, J., Xiao, S., Zhang, P., Luo, K., Lian, D., and Liu, Z.: Bge m3-embedding: Multi-lingual, multi-functionality, multi-granularity text embeddings through self-knowledge distillation, arXiv preprint arXiv:2402.03216, 4, 2024.

Chen, T., Kornblith, S., Norouzi, M., and Hinton, G.: A Simple Framework for Contrastive Learning of Visual Representations, arXiv preprint arXiv:2002.05709, 2020.

Chen, X. and He, K.: Exploring Simple Siamese Representation Learning, arXiv preprint arXiv:2011.10566, 2020.

Cong, Y., Khanna, S., Meng, C., Liu, P., Rozi, E., He, Y., Burke, M., Lobell, D., and Ermon, S.: Satmae: Pre-training transformers for temporal and multi-spectral satellite imagery, Advances in Neural Information Processing Systems, 35, 197–211, 2022.

Corbane, C., Syrris, V., Sabo, F., Politis, P., Melchiorri, M., Pesaresi, M., Soille, P., and Kemper, T.: Convolutional neural networks for global human settlements mapping from Sentinel-2 satellite imagery, Neural Computing and Applications, 33, 6697–6720, 2021.

Elvidge, C. D., Baugh, K., Zhizhin, M., Hsu, F. C., and Ghosh, T.: VIIRS night-time lights, International journal of remote sensing, 38, 5860–5879, 2017.

Fedus, W., Zoph, B., and Shazeer, N.: Switch transformers: Scaling to trillion parameter models with simple and efficient sparsity, Journal of Machine Learning Research, 23, 1–39, 2022.

Feng, Z., Jaffer, S., Knezevic, J., Sormunen, S., Young, R., Lisaius, M., Immitzer, M., Ball, J., Atzberger, C., and Coomes, D. A.: TESSERA: Temporal Embeddings of Surface Spectra for Earth Representation and Analysis, arXiv e-prints, arXiv: 2506.20380, 2025.

Grill, J.-B., Strub, F., Altché, F., Tallec, C., Richemond, P. H., Buchatskaya, E., Doersch, C., Pires, B. A., Guo, Z. D., and Azar, M. G.: Bootstrap your own latent: A new approach to self-supervised Learning, arXiv preprint arXiv:2006.07733, 2020.

Haklay, M. and Weber, P.: Openstreetmap: User-generated street maps, IEEE Pervasive computing, 7, 12–18, 2008.

He, K., Fan, H., Wu, Y., Xie, S., and Girshick, R.: Momentum Contrast for Unsupervised Visual Representation Learning, arXiv e-prints, arXiv: 1911.05722, 2019.

Hou, S., Jiao, H., Liang, J., Shen, Z., Zhao, A., and Wu, H.: GeoCogent: an LLM-based agent for geospatial code generation, International Journal of Geographical Information Science, 40, 1073–1106, 2026a.

Hou, S., Jiao, H., Liu, Z., Xie, L., Chen, G., Wu, S., Guan, X., and Wu, H.: GeoSQL-Eval: first evaluation of LLMs on PostGIS-based NL2GeoSQL queries, Expert Systems with Applications, 132122, 2026b.

Hou, S., Jiao, H., Shen, Z., Liang, J., Zhao, A., Zhang, X., Wang, J., and Wu, H.: Chain-of-programming (CoP): empowering large language models for geospatial code generation task, International Journal of Digital Earth, 18, 2509812, 2025a.

Hou, S., Shen, Z., Zhao, A., Liang, J., Gui, Z., Guan, X., Li, R., and Wu, H.: GeoCode-GPT:

A large language model for geospatial code generation, International Journal of Applied Earth Observation and Geoinformation, 138, 104456, 2025b.
Hou, S., Jiao, H., Xu, Q., Qing, Y., Liu, Z., Xie, L., Xu, Z., Guan, X., and Wu, H.: Can AlphaEarth Foundations Redefine the Paradigm of Gridded Population Mapping? A Systematic Evaluation across 18 Global Cities and Large-Scale Mapping Applications, 2026c.
Hou, S., Jiao, H., Liu, Z., Xie, L., Chen, G., Wu, S., Xu, Z., Wang, Z., Tang, S., and Qing, Y.: AlphaEarth Foundations (AEF) in Earth Observation: A Systematic Review of Applications and Practices, 2026d.
Hou, S., Liu, Z., Jiao, H., Xu, Z., Zhang, X., Xie, L., Qing, Y., Liang, J., Guan, X., and Wu, H.: Slum Detection and Density Mapping with AlphaEarth Foundations: A Representation Learning Evaluation Across 12 Global Cities, arXiv preprint arXiv:2605.10029, 2026e.
Hu, J., Shen, L., Albanie, S., Sun, G., and Wu, E.: Squeeze-and-Excitation Networks, arXiv e-prints, 2018.
Huang, X., Song, Y., Yang, J., Wang, W., Ren, H., Dong, M., Feng, Y., Yin, H., and Li, J.: Toward accurate mapping of 30-m time-series global impervious surface area (GISA), International Journal of Applied Earth Observation and Geoinformation, 109, 102787, 2022.
Jakubik, J., Roy, S., Phillips, C., Fraccaro, P., Godwin, D., Zadrozny, B., Szwarcman, D., Gomes, C., Nyirjesy, G., and Edwards, B.: Foundation models for generalist geospatial artificial intelligence, arXiv preprint arXiv:2310.18660, 2023.
Jean, N., Burke, M., Xie, M., Alampay Davis, W. M., Lobell, D. B., and Ermon, S.: Combining satellite imagery and machine learning to predict poverty, Science, 353, 790–794, 2016.
Klemmer, K., Rolf, E., Robinson, C., Mackey, L., and Rußwurm, M.: SatCLIP: Global, General-Purpose Location Embeddings with Satellite Imagery, arXiv preprint arXiv:2311.17179, 2023.
Liu, J., Qin, Q., Dong, G., Wang, X., Feng, J., Zeng, Z., and Cheng, T.: Beyond AlphaEarth: toward human-centered spatial representation via POI-guided contrastive learning, arXiv preprint arXiv:2510.09894, 2025.
Lloyd, C. T., Chamberlain, H., Kerr, D., Yetman, G., Pistolesi, L., Stevens, F. R., Gaughan, A. E., Nieves, J. J., Hornby, G., and MacManus, K.: Global spatio-temporally harmonised datasets for producing high-resolution gridded population distribution datasets, Big earth data, 3, 108–139, 2019.
McInnes, L., Healy, J., and Melville, J.: Umap: Uniform manifold approximation and projection for dimension reduction, arXiv preprint arXiv:1802.03426, 2018.
Nazir, U., Cheng, I.-H., and Khalid, S.: AlphaEarth Satellite Embeddings for Modelling Climate Sensitive Diseases Towards Global Health Resilience, arXiv preprint arXiv:2605.10949, 2026.
Pedregosa, F., Varoquaux, G., Gramfort, A., Michel, V., Thirion, B., Grisel, O., Blondel, M., Müller, A., Nothman, J., and Louppe, G.: Scikit-learn: Machine Learning in Python, arXiv e-prints, 1201, 2012.
Pesaresi, M. and Politis, P.: GHS-BUILT-H R2023A-GHS building height, derived from AW3D30, SRTM30, and Sentinel2 composite (2018), European Commission, Joint Research Centre (JRC), 2023.
Pettersson, M. B. and Daoud, A.: Leveraging Compact Satellite Embeddings and Graph Neural Networks for Large-Scale Poverty Mapping, arXiv preprint arXiv:2511.01408, 2025.
Radford, A., Kim, J. W., Hallacy, C., Ramesh, A., Goh, G., Agarwal, S., Sastry, G., Askell, A., Mishkin, P., and Clark, J.: Learning Transferable Visual Models From Natural Language Supervision, arXiv preprint arXiv:2103.00020, 2021.
Reed, C. J., Gupta, R., Li, S., Brockman, S., Funk, C., Clipp, B., Keutzer, K., Candido, S., Uyttendaele, M., and Darrell, T.: Scale-MAE: A Scale-Aware Masked Autoencoder for Multiscale Geospatial Representation Learning, arXiv preprint arXiv:2212.14532, 2022.
Roberts, D. R., Bahn, V., Ciuti, S., Boyce, M. S., Elith, J., Guillera-Arroita, G., Hauenstein, S., Lahoz-Monfort, J. J., Schröder, B., and Thuiller, W.: Cross-validation strategies for data with temporal, spatial, hierarchical, or phylogenetic structure, Ecography, 40, 913–929, 2017.
Rolf, E., Proctor, J., Carleton, T., Bolliger, I., Shankar, V., Ishihara, M., Recht, B., and Hsiang, S.: A generalizable and accessible approach to machine learning with global satellite imagery,

Nature communications, 12, 4392, 2021.
Shazeer, N., Mirhoseini, A., Maziarz, K., Davis, A., Le, Q., Hinton, G., and Dean, J.: Outrageously large neural networks: The sparsely-gated mixture-of-experts layer, arXiv preprint arXiv:1701.06538, 2017.
Tobler, W. R.: A computer movie simulating urban growth in the Detroit region, Economic geography, 46, 234–240, 1970.
Vaswani, A., Shazeer, N., Parmar, N., Uszkoreit, J., Jones, L., Gomez, A. N., Kaiser, L., and Polosukhin, I.: Attention Is All You Need, arXiv preprint arXiv:1706.03762, 2017.
Wan, Z., Hook, S., and Hulley, G.: MOD11A2 MODIS/Terra land surface temperature/emissivity 8-day L3 global 1km SIN grid V006, NASA EOSDIS Land Processes Distributed Active Archive Center (DAAC) data set, 2015.
Xiong, Z., Wang, Y., Zhang, F., Stewart, A. J., Hanna, J., Borth, D., Papoutsis, I., Saux, B. L., Camps-Valls, G., and Zhu, X. X.: Neural plasticity-inspired multimodal foundation model for earth observation, arXiv preprint arXiv:2403.15356, 2024.
Yang, J. and Huang, X.: The 30 m annual land cover dataset and its dynamics in China from 1990 to 2019, Earth System Science Data, 13, 3907–3925, 2021.
Zbontar, J., Jing, L., Misra, I., LeCun, Y., and Deny, S.: Barlow Twins: Self-Supervised Learning via Redundancy Reduction, arXiv preprint arXiv:2103.03230, 2021.

**Appendix A**

Appendix A presents the list of 36 validation cities used in this study, including their geographic regions, city codes, socioeconomic tiers, WGS84 bounding box coordinates, and estimated physical dimensions of each bounding box.

**Table A1. Geographic distribution, city tiers, and bounding box definitions for the 36 validation cities.**

| Region | Code | City | Tier | Lon_min | Lat_min | Lon_max | Lat_max | Width km | Height km |
|---|---|---|---|---|---|---|---|---|---|
| Central China | CGO | Zhengzhou | T2 | 113.45 | 34.60 | 113.80 | 34.90 | 32.00 | 33.20 |
| | CSX | Changsha | T1.5 | 112.85 | 28.10 | 113.10 | 28.30 | 24.50 | 22.10 |
| | WUH | Wuhan | T1.5 | 114.18 | 30.45 | 114.50 | 30.75 | 30.70 | 33.20 |
| East China | FOC | Fuzhou | T3 | 119.15 | 25.95 | 119.45 | 26.20 | 30.00 | 27.60 |
| | KHN | Nanchang | T3 | 115.75 | 28.55 | 116.05 | 28.80 | 29.30 | 27.60 |
| | HFE | Hefei | T2 | 117.05 | 31.70 | 117.40 | 31.95 | 33.10 | 27.60 |
| | HGH | Hangzhou | T1.5 | 119.95 | 30.10 | 120.35 | 30.40 | 38.50 | 33.20 |
| | NGB | Ningbo | T1.5 | 121.35 | 29.75 | 121.75 | 30.05 | 38.60 | 33.20 |
| | NKG | Nanjing | T1.5 | 118.60 | 31.90 | 118.95 | 32.20 | 33.00 | 33.20 |
| | SHA | Shanghai | T1 | 121.20 | 31.05 | 121.65 | 31.42 | 42.80 | 40.90 |
| | TAO | Qingdao | T2 | 120.20 | 35.90 | 120.55 | 36.20 | 31.50 | 33.20 |
| | TNA | Jinan | T2 | 116.80 | 36.55 | 117.20 | 36.80 | 35.70 | 27.60 |
| | XMN | Xiamen | T1.5 | 117.95 | 24.40 | 118.25 | 24.65 | 30.40 | 27.60 |
| North China | HET | Hohhot | T2 | 111.60 | 40.70 | 111.90 | 41.00 | 25.30 | 33.20 |
| | PEK | Beijing | T1 | 116.15 | 39.75 | 116.60 | 40.10 | 38.40 | 38.70 |
| | SJW | Shijiazhuang | T2 | 114.35 | 37.90 | 114.70 | 38.20 | 30.70 | 33.20 |
| | TSN | Tianjin | T1 | 117.00 | 38.95 | 117.45 | 39.30 | 38.90 | 38.70 |
| | TYN | Taiyuan | T2 | 112.40 | 37.75 | 112.70 | 38.00 | 26.40 | 27.60 |
| Northeast China | CGQ | Changchun | T2 | 125.15 | 43.70 | 125.50 | 44.00 | 28.10 | 33.20 |
| | DLC | Dalian | T2 | 121.45 | 38.80 | 121.80 | 39.10 | 30.30 | 33.20 |
| | HRB | Harbin | T2 | 126.35 | 45.65 | 126.75 | 46.00 | 31.00 | 38.70 |
| | SHE | Shenyang | T2 | 123.25 | 41.65 | 123.65 | 41.95 | 33.20 | 33.20 |
| Northwest China | INC | Yinchuan | T2 | 106.10 | 38.40 | 106.30 | 38.55 | 17.40 | 16.60 |
| | LHW | Lanzhou | T3 | 103.70 | 35.95 | 104.00 | 36.20 | 27.00 | 27.60 |
| | SIA | Xi’an | T2 | 108.85 | 34.15 | 109.10 | 34.35 | 23.00 | 22.10 |
| | URC | Urumqi | T3 | 87.45 | 43.70 | 87.80 | 43.95 | 28.10 | 27.60 |
| | XNN | Xining | T3 | 101.65 | 36.50 | 101.95 | 36.75 | 26.80 | 27.60 |
| South China | CAN | Guangzhou | T1 | 113.05 | 22.95 | 113.50 | 23.30 | 46.10 | 38.70 |
| | HAK | Haikou | T2 | 110.20 | 19.95 | 110.50 | 20.15 | 31.40 | 22.10 |
| | NNG | Nanning | T2 | 108.20 | 22.65 | 108.55 | 22.95 | 35.90 | 33.20 |
| | SZX | Shenzhen | T1 | 113.80 | 22.45 | 114.30 | 22.75 | 51.40 | 33.20 |
| Southwest China | CKG | Chongqing | T1 | 106.35 | 29.40 | 106.75 | 29.75 | 38.70 | 38.70 |
| | CTU | Chengdu | T1.5 | 103.95 | 30.55 | 104.20 | 30.78 | 23.90 | 25.40 |
| | KMG | Kunming | T2 | 102.65 | 24.90 | 103.00 | 25.20 | 35.30 | 33.20 |
| | KWE | Guiyang | T2 | 106.55 | 26.50 | 106.85 | 26.80 | 29.80 | 33.20 |
| | LXA | Lhasa | T4 | 91.00 | 29.55 | 91.30 | 29.75 | 29.00 | 22.10 |

## Appendix B

This appendix details the acquisition, processing, and rasterization procedures for the seven categories of training input data streams (T1–T7).

### B.1 T1—AlphaEarth Foundations Embeddings

The T1 data stream consists of AlphaEarth Foundations embeddings derived from the Google Earth Engine dataset GOOGLE/SATELLITE_EMBEDDING/V1/ANNUAL (Brown et al., 2025). This product provides annual dense embeddings at a global spatial resolution of 10 m, represented as 64-byte vectors with 8-bit quantization, and covers the period from 2017 to 2024. In this study, Export.image.toDrive tasks were submitted through the GEE Code Editor to export the corresponding raster data sequentially by year and by the city bounding boxes defined in Appendix A. Owing to the output-capacity limits of individual GEE tasks, including EECU quotas and the 16 GB shard limit, large cities such as Shanghai and Beijing required additional tiled exports under multiple resolution settings. The tiled files were named using the convention {CODE}_AEF_{year}-NNNN-NNNN.tif, where the last two numeric segments indicate the pixel offsets of the tile within the original image. During preprocessing, the 64-band AEF rasters were used directly as training inputs without inverse quantization, because AEF is designed to use quantized values in downstream linear-probe tasks. For pixel-center queries, the corresponding 64-dimensional embedding vector was extracted from the longitude–latitude coordinate (lng,lat) using rasterio.sample.

### B.2 T2—WorldPop Constrained Population

The T2 data stream consists of the WorldPop Constrained Population product released by the University of Southampton (Lloyd et al., 2019), which provides annual population rasters at a spatial resolution of 100 m. The data were acquired in two ways: first, by exporting annual rasters from the GEE dataset WorldPop/GP/100 m/pop; and second, by downloading country-level files from the official WorldPop FTP server and clipping them to the study-city boundaries. During preprocessing, this stream was treated as a single-band continuous variable representing the total population within each pixel. Because population distributions are strongly heavy-tailed, a log1p transformation was applied before training to compress extreme values and improve the stability of the numerical distribution.

### B.3 T3—Nighttime Light Statistics

The T3 data stream comprises nighttime light statistical features derived from the NOAA VIIRS Day/Night Band Monthly Composites, including VCMSLCFG and VCMCFG products, as well as the NOAA DMSP-OLS Stable Lights 2013 benchmark data. Nighttime light data were obtained through the GEE datasets NOAA/VIIRS/DNB/MONTHLY_V1/VCMSLCFG, NOAA/VIIRS/DNB/MONTHLY_V1/VCMCFG, and NOAA/DMSP-OLS/NIGHTTIME_LIGHTS, and were composited by year. The native spatial resolution of these products is 500 m. During preprocessing for the 36-city benchmark training, the 18-dimensional nighttime light statistical rasters were clipped to the city bounding boxes and uniformly resampled to 100 m to match the spatial scale of the other training input streams. Before training, each band was then standardized using z-score normalization based on the mean and standard deviation computed over the full dataset. This procedure reduced scale differences among nighttime light statistics and improved the stability of model training. The composition of the 18 bands is provided in Table B1.

**Table B1. Composition and definitions of the 18-dimensional nighttime light statistical features**

| Band | Name | Description |
|---|---|---|
| 1 | viirs_mean | Mean monthly radiance over the year |
| 2 | viirs_median | Median monthly radiance over the year |
| 3 | viirs_max | Maximum monthly radiance over the year |
| 4 | viirs_min | Minimum monthly radiance over the year |
| 5 | viirs_std | Standard deviation of monthly radiance over the year |
| 6 | viirs_p10 | 10th percentile of monthly radiance over the year |
| 7 | viirs_p90 | 90th percentile of monthly radiance over the year |

| Band | Name | Description |
|---|---|---|
| 8 | viirs_cv | Coefficient of variation, defined as std / (mean + 1e-6) |
| 9 | viirs_lit_fraction | Fraction of months with radiance greater than 0 over the year |
| 10 | viirs_summer | Mean radiance from June to August |
| 11 | viirs_winter | Mean radiance for December and the following January–February |
| 12 | viirs_seasonal_diff | summer−winter |
| 13 | viirs_focal_mean | Mean radiance within a 3 × 3 pixel neighborhood |
| 14 | viirs_local_contrast | mean−focal_mean |
| 15 | viirs_sl_mean | Annual mean radiance from VCMCFG, including stray-light contamination |
| 16 | viirs_straylight | Stray-light component, defined as sl_mean − mean |
| 17 | dmsp_2013_baseline | DMSP-OLS 2013 stable-light baseline |
| 18 | ntl_growth_vs_dmsp | Relative growth indicator against 2013, defined as viirs_mean / 27.65 − dmsp_2013 / 63.0 |

### B.4 T4—Remote Sensing Indices

The T4 data stream consists of remote sensing index features derived from the USGS Landsat 8 Collection 2 Tier 1 Level-2 dataset (LANDSAT/LC08/C02/T1_L2) and the JRC GHSL P2023A GHS-BUILT-S dataset (JRC/GHSL/P2023A/GHS_BUILT_S). The relevant data were exported annually through GEE. For Landsat 8 imagery, scenes were first preliminarily filtered using CLOUD_COVER < 30%, and cloud and cloud-shadow pixels were removed using the QA_PIXEL quality-control band by masking Bit 3 and Bit 4. Surface reflectance was then restored according to the scaling rule of the Collection 2 Level-2 product, using a scale factor of 0.0000275 and an additive offset of -0.2. For the GHSL GHS-BUILT-S data, the 2015 and 2020 bi-temporal products were used as the basis. Built-surface information for intermediate and extrapolated years was generated through linear interpolation, and pixel values were normalized to the range of [0, 1] using built_surface / 10000. During preprocessing, all remote sensing index features were output at a unified spatial resolution of 100 m and spatially aligned with the grid of the T2 WorldPop population data. The composition of the three bands is provided in Table B2.

**Table B2. Composition of the 3-dimensional feature bands for the T4 data stream**

| Band | Name | Description |
|---|---|---|
| 1 | ndvi | Normalized Difference Vegetation Index, defined as (NIR−Red)/(NIR+Red) |
| 2 | ndbi | Normalized Difference Built-up Index, defined as (SWIR1−NIR)/(SWIR1+NIR) |
| 3 | ghsl_built | Proportion of GHSL built-up surface, normalized to the range of [0, 1] |

### B.5 T5—Point-of-Interest Raster Features

The T5 data stream consists of point-of-interest (POI) raster features derived from the Amap Open Platform (amap.com) Place Search API. We collected complete POI records for the 36 study cities from 2017 to 2024 and subsequently extended the collection to 324 cities nationwide. The raw POI fields included name, address, longitude and latitude, category code, and detailed subclass. The category codes followed Amap's official 16-class classification system. During training, four categories—place_name, residential, living_service, and auto_service—were excluded. In the rasterization stage, each 100 m pixel center was treated as a query location, and nearby POIs were retrieved using a KDTree. Two sets of 12-dimensional features were then constructed. The first set consists of density features, defined as the number of POIs in each category within a 500 m radius divided by 0.785 km$^2$, representing POI density per unit area. The second set consists of distance features, defined as the distance from the pixel center to the nearest POI of the corresponding category, measured in kilometers. If no POI of a given category was present, the corresponding density was set to 0, and the distance was assigned a fallback value of 5 km. During preprocessing, a log1p transformation was applied to the density stream to compress its heavy-tailed distribution, whereas the distance stream was retained in its original scale; both streams were subsequently standardized by z-score normalization before training. In the nationwide 324-city stage, some cities had missing POI collections in 2017 and 2022. These gaps were filled using the union of adjacent years: if both preceding and following years were available, their union was used; if only one adjacent year was available, that year was directly reused. This strategy achieved 100% POI coverage for all 324 cities across the eight-year period. To accommodate the model input requirements and the representation of urban functions, we consolidated Amap's 16 POI categories into 12 study-specific categories. The detailed mapping is provided in Table B3.

**Table B3. Category composition and source mapping of the 12 POI classes for the T5 POI raster features.**

| 12-category ID | 12-category name | Source category in Amap's 16-class system | Notes |
|---|---|---|---|
| 1 | food | food | Food and catering |
| 2 | education | education_culture | Education and culture |
| 3 | healthcare | healthcare | Healthcare |
| 4 | finance | finance | Finance |
| 5 | shopping | shopping | Shopping |
| 6 | entertainment | leisure_sport | Entertainment and sports, including KTV venues, cinemas, bars, and sports facilities |
| 7 | public_service | government+infrastructure | Government agencies and public facilities |
| 8 | leisure | tourism | Tourist attractions and scenic sites |
| 9 | accommodation | accommodation | Accommodation |
| 10 | religion | Keywords in the name field | Religious sites identified from keywords in the name field, including temples, churches, mosques, Taoist temples, Zen monasteries, monasteries, shrines, and Buddhist halls |
| 11 | transport_service | transport | Transportation services |
| 12 | office | company | Corporate and office facilities |

### B.6 T6—Urban Morphological and Spatial Attributes

The T6 data stream consists of urban morphological and spatial attributes derived from OpenStreetMap Planet History .osh.pbf data obtained from Geofabrik and planet.openstreetmap.org. This dataset contains historical global vector information, including road networks, water bodies, railways, and public transit features. We used osmium-tool to clip the data sequentially by city bounding box and year. Specifically, for each city, osmium extract --strategy complete_ways --history was first used to extract the historical PBF file for the corresponding spatial extent; osmium time-filter was then applied to generate annual OSM snapshot data for each city. During rasterization, each 100 m pixel center was treated as a query location, and a spatial index was used to extract nine urban morphological and spatial features, which were then output as 100 m resolution rasters. During preprocessing, distance-related features were retained in their original scale, whereas density- and length-related features were transformed using log1p to compress heavy-tailed distributions. All features were subsequently standardized using z-score normalization before training. The T6 data stream constructs nine-dimensional urban morphological spatial attributes based on OSM historical snapshot data, characterizing pixel-level accessibility to urban centers, proximity to blue–green spaces, road network density, and public transit accessibility. The detailed feature composition is provided in Table B4.

**Table B4. Composition of the nine-dimensional spatial structure features for the T6 urban morphology data stream.**

| Dimension | Name | Description |
|---|---|---|
| 1 | dist_center_main | Euclidean distance to the main urban center, defined as the provincial/municipal government location or the core urban area, measured in kilometers |
| 2 | dist_water | Distance to the nearest water body, including rivers, lakes, and canals, measured in kilometers |
| 3 | road_arterial | Total length of arterial roads within a 1 km radius, including motorway, trunk, and primary roads, measured in kilometers |
| 4 | road_collector | Total length of collector roads within a 1 km radius, including secondary and tertiary roads, measured in kilometers |
| 5 | road_local | Total length of local and residential-access roads within a 1 km radius, including residential and service roads, measured in kilometers |
| 6 | road_density | Road density, calculated as the total length of all roads within a 1 km radius divided by 3.14, with units of km/km² |
| 7 | dist_metro | Distance to the nearest metro station (railway=subway), measured in kilometers; for cities without metro systems, a fallback value of 30 km was assigned |
| 8 | metro_density | Number of metro stations within a 1 km radius |
| 9 | dist_bus | Distance to the nearest bus stop or bus station (highway=bus_stop or amenity=bus_station), measured in kilometers |

### B.7 T7—Cross-lingual POI Text Embeddings

The T7 data stream consists of cross-lingual POI text embeddings. Its raw data share the same source as T5, namely the Amap POI database, and are further processed using large language models, either OpenAI GPT-4 or DeepSeek-V3, together with the BGE-M3 cross-lingual embedding model (Chen et al., 2024), for semantic abstraction and vectorized representation. The construction procedure comprises three steps. First, a structured POI list is extracted. For each pixel, all POIs within a 200 m buffer are retrieved and concatenated into a structured text list ($T_{raw}$) in the format "category–name". For example: "food–Old Beijing Zhajiangmian Restaurant; food–Starbucks (Guomao Branch); shopping–Guomao Phase III Luxury Retail; office–COFCO Plaza Office Building; finance–ICBC Guomao Branch; transport_service–Guomao Metro Station." Second, LLM-based abstraction is performed. ($T_{raw}$) is provided to the large language model through a structured prompt (P), which instructs the model to generate a de-identified functional-zone description ($T_{abs}$). The prompt requires the model to preserve urban tier information (T1/T1.5/T2/T3/T4), administrative location context (CBD/near-suburban/suburban/old urban core), the composition of functional categories, and business-grade information (high-end/mid-range/budget), while prohibiting the retention of any specific brand names, institution names, or geographic proper names, such as "Starbucks," "Hualian," "COFCO Plaza," "Guomao," and "Lujiazui." The output is a natural-language paragraph of 100–200 Chinese characters, for example: "This area is located in the core business district of a first-tier city, dominated by mid- to high-end catering, luxury retail, Grade-A office buildings, and financial services. It is close to metro stations and surrounded by dense public service facilities, and can be characterized overall as a high-end mixed business functional zone." Third, BGE-M3 encoding is conducted. ($T_{abs}$) is input into the BGE-M3 model (BAAI/bge-m3) and encoded as a 1024-dimensional dense vector ($e_{abs}$). BGE-M3 is a general-purpose multilingual embedding model that supports more than 100 languages, including Chinese and English, and performs strongly in cross-lingual semantic retrieval tasks. As an ablation baseline, the unmodified raw text ($T_{raw}$) is also directly encoded by BGE-M3 to obtain the raw text embedding ($e_{raw}$). In the experiments, $e_{abs}$ is used as the text input, and five configurations, TE1–TE5 (§4.1.5), are employed to compare the effects of different alignment strategies and weights on cross-city generalization performance. $e_{raw}$ is released together with the dataset for comparison in future studies, although a systematic ablation between $e_{abs}$ and $e_{raw}$ is not conducted in this study. During nationwide inference, use_text=False is set, and T7 is not involved in the encoder forward pass. The semantic information provided by text alignment has already been internalized into the encoder weights during training; therefore, generating T7 text embeddings is unnecessary for inference over 324 cities across eight years, substantially reducing the deployment cost of the product.

**B.8 Raster Alignment of Data Streams**

Before being fed into the encoder, all six numerical data streams (T1–T6) were resampled to a unified 100 m WGS84 grid and spatially aligned with the study bounding box of each of the 36 cities. Specifically, the T1 AlphaEarth Foundations embeddings, originally provided at 10 m resolution, were downsampled to 100 m using bilinear interpolation; the T3 nighttime light data, originally provided at 500 m resolution, were upsampled to 100 m using nearest-neighbor interpolation; and T2, T4, T5, and T6, all originally available at 100 m resolution, were retained at their native spatial resolution. For each pixel location (city,year,lng,lat), a 119-dimensional numerical feature vector from T1–T6 was extracted using rasterio.sample and used as the unified numerical input to the encoder.

**B.9 Standardization of Data Streams**

All input streams were standardized dimension-wise using z-score normalization, with the mean and standard deviation computed from the 36-city benchmark dataset. Specifically, for any dimension d, the standardization procedure is defined as follows:

$$x_{normalized}[d] = \frac{x[d] - \mu[d]}{\sigma[d] + \epsilon} \quad (B1)$$

Here, $\mu[d]$ and $\sigma[d]$ denote the mean and standard deviation of the corresponding dimension computed over all 14,400,000 pixel–year samples, respectively, and $\epsilon = 10^{-6}$ is used to prevent division by zero. All standardization parameters were stored in samples/meta.json and directly reused during nationwide inference to ensure consistency between the input distributions during inference and training.

**Appendix C**

This appendix details the acquisition, processing, and construction procedures for the 16 downstream validation labels (Vd1–Vd16). These labels are organized into two categories: physical-Observable labels (PO, Vd1–Vd7) and socioeconomic -anchored labels (SEA, Vd8–Vd16).

**C.1 Physical-Observable Labels (PO, Vd1–Vd7)**

**Vd1—Normalized Difference Vegetation Index (NDVI)**

The Vd1 label corresponds to the NDVI. Its data source is the same as that used for the T4 training stream, namely Landsat 8 Collection 2 Tier 1 Level-2. The construction procedure follows §B.4: annual image composites were generated, and cloud and cloud-shadow pixels were removed using CLOUD_COVER < 30% and the QA_PIXEL cloud mask. NDVI was then calculated from the near-infrared band SR_B5 and the red band SR_B4 as follows:

$$NDVI = \frac{NIR - Red}{NIR + Red} \quad (C1)$$

The final output was a raster at 100 m spatial resolution. In this study, this raster was used as a downstream validation label; before training, the label values were divided by 10,000 to restore them to the range of [-1, 1].

**Vd2—Normalized Difference Built-up Index (NDBI)**

The Vd2 label corresponds to the NDBI. Its data source is the same as that used for Vd1, and its construction procedure follows §B.4. This label was calculated from the short-wave infrared band (SWIR1) and the near-infrared band (NIR) as follows:

$$NDBI = \frac{SWIR1 - NIR}{SWIR1 + NIR} \quad (C2)$$

After calculation, the result was output as a 100 m raster. Before training, the label values were divided by 10,000 for numerical restoration.

**Vd3—Proportion of GHSL Built-up Surface**

The Vd3 label represents the proportion of GHSL built-up surface. The data source is the JRC GHSL P2023A GHS-BUILT-S product, which is the same source as the third dimension of the T4 training stream. Its construction procedure follows §B.4: the 2015 and 2020 bi-temporal products were used as the basis, values for intermediate years were generated through linear interpolation, and pixel values were divided by 10,000 to normalize them to the range of [0, 1]. As this label is already provided as a normalized raster, no additional transformation was applied before training.

**Vd4—CLCD land-use classification**

The Vd4 label corresponds to CLCD land-use classification, derived from the China Land Cover Dataset developed by Wuhan University (Yang and Huang, 2021). This dataset is an annual land-cover product at 30 m spatial resolution and contains nine integer classes: 1 cropland, 2 forest, 3 grassland, 4 shrubland, 5 water, 6 snow/ice, 7 barren land, 8 impervious surface, and 9 wetland.

We downloaded the nationwide rasters from Zenodo, clipped them to the bounding boxes of the 36 cities, and resampled them to 100 m using nearest-neighbor interpolation to preserve the integer class labels. Vd4 was treated as a classification task and evaluated using macro-averaged F1 score ($F1_{\mathrm{macro}}$) and Cohen's $\kappa$.

**Vd5—MODIS Land Surface Temperature**

The Vd5 label represents land surface temperature (LST), derived from the MOD11A2 V6.1 product (MODIS/061/MOD11A2) (Wan et al., 2015). This product provides 8-day composite LST data at a spatial resolution of 1000 m. Using GEE, we extracted the annual median of LST_Day_1km for each year, converted it to Kelvin using the MODIS standard scale factor of 0.02, and then converted it to degrees Celsius by subtracting 273.15. The resulting raster was subsequently resampled to 100 m resolution. This label is expressed in °C and was used directly before training without any additional transformation.

**Vd6—Building Height**

The Vd6 label represents building height, derived from the JRC GHSL P2023A GHS-BUILT-H 2018 product (JRC/GHSL/P2023A/GHS_BUILT_H/2018) (Pesaresi and Politis, 2023). This dataset provides a single-temporal building-height raster at 100 m resolution, with units of meters. The raster was clipped to the bounding boxes of the 36 cities, and no temporal extrapolation was performed. Therefore, Vd6 uses the same static 2018 building-height data for all years from 2017 to 2024. Before training, a log1p transformation was applied to this label to compress its right-skewed distribution and the large number of zero values.

**Vd7—Built Surface Area**

The Vd7 label represents built surface area, derived from the JRC GHSL P2023A GHS-BUILT-S product, which is the same source as the third dimension of the T4 training stream. However, the two variables differ in definition: T4 uses the normalized built_fraction, whereas Vd7 uses the original built_surface values, with units of $m^2$ per 100 m × 100 m pixel. The GHSL bi-temporal processing and linear interpolation procedure followed §B.4. Before training, a log1p transformation was applied to this label.

**C.2 Socioeconomic-Anchored Labels (SEA, Vd8–Vd16)**

**Vd8—Population Density**

The Vd8 label represents population density, derived from the WorldPop Constrained Population product, which is identical to the data source used for the T2 training stream. Its construction procedure follows §B.2. The label is a single-band 100 m raster with units of persons per pixel. Before training, a log1p transformation was applied. Because this label shares the same source as the T2 input stream, it serves as a test of knowledge internalization, assessing whether the population signal in T2 has been successfully injected into the embedding product.

**Vd9—Mean Nighttime Light Intensity**

The Vd9 label represents mean nighttime light intensity, derived from the NOAA VIIRS DNB Monthly VCMSLCFG product. It is the same source as Band 1 of the T3 training stream, viirs_mean, which represents the mean monthly radiance over the year. This dataset is a single-band raster at 500 m resolution and was resampled to 100 m before being used as a validation label. A log1p transformation was applied before training. This label also serves as a test of knowledge internalization.

**Vd10—18-dimensional Nighttime Light Statistics**

The Vd10 label consists of 18-dimensional nighttime light statistical features and shares the same data source as the T3 training stream. The composition of its 18 bands is described in §B.3. Different pre-training transformations were applied to each dimension according to its physical meaning. Intensity-related dimensions, such as viirs_mean, viirs_median, viirs_max, viirs_p10, viirs_p90, and viirs_focal_mean, were transformed using log1p to compress heavy-tailed distributions. Difference-related dimensions that can take either positive or negative values, such as viirs_local_contrast, viirs_seasonal_diff, viirs_straylight, and ntl_growth_vs_dmsp, were transformed using signed_log1p, defined as follows:

$$signed_log1p(x) = sign(x) \cdot log(1 + |x|) \qquad (C3)$$

This is transformation preserves sign information. Ratio-based dimensions that were already normalized, such as viirs_cv and viirs_lit_fraction, were used directly. This label also serves as a test of knowledge internalization and is used to evaluate the completeness with which the 18-dimensional nighttime light statistics are retained in the embeddings.

**Vd11—POI Density (12 Categories)**

The Vd11 label represents POI density and is derived from the Amap Open Platform POI database, the same source as the T5 training stream. Its construction procedure follows §B.5. Specifically, for each pixel, the number of POIs in each of the 12 categories within a 500 m radius was counted and divided by 0.785 km² to obtain POI density per unit area, with units of counts per km². The label was output as a 12-dimensional raster, and a log1p transformation was applied to each dimension before training.

**Vd12—POI Distance (12 Categories)**

The Vd12 label represents POI distance and uses the same data source as Vd11. Its construction procedure follows the distance component described in §B.5. For each pixel, the Euclidean distance to the nearest POI of each corresponding category was calculated in kilometers. If no POI of a given category was available, a fallback value of 5 km was assigned. The label was output as a 12-dimensional raster and used directly before training without additional transformation.

**Vd13—Urban Morphology (9 Dimensions)**

The Vd13 label consists of nine-dimensional urban morphological and spatial-structure features derived from OpenStreetMap, the same source as the T6 training stream. The composition and construction procedure of these nine dimensions are detailed in §B.6. Before training, the dist_center_main dimension was used directly, whereas the remaining eight dimensions—including distance to water bodies, three-level road-network lengths, road density, metro distance, metro density, and bus-stop distance—were transformed using log1p.

**Vd14—First Built-up Year of GISA Impervious Surface**

The Vd14 label represents the first built-up year of impervious surfaces from the GISA global impervious surface dynamics dataset developed by Wuhan University (Huang et al., 2022). This dataset provides a 30 m global raster of the first year in which each pixel became impervious, with values ranging from 1985 to 2021; a value of 0 indicates that the pixel had not become impervious by 2021. We clipped the dataset to the bounding boxes of the 36 cities and resampled it to 100 m using nearest-neighbor interpolation to preserve the integer year values. In addition, an auxiliary binary field, is_built_by_2020, was generated to indicate whether each pixel had become impervious before 2020. Vd14 was treated as a regression task, and the first built-up year, expressed in calendar years, was used directly before training without further transformation.

**Vd15—Intra-city Housing Prices**

The Vd15 label represents intra-city housing prices, derived from community-level second-hand residential listing records continuously tracked and compiled by our research team from 2017 to 2024. The raw data were collected from publicly visible listings on major real-estate information platforms. The data collection was conducted through periodic manual queries by the research team, recording publicly available fields such as average listing price, housing layout, and year of construction for each residential community. The data were used solely for academic research and did not involve commercial redistribution or large-scale automated acquisition from the platforms. The raw dataset covers the 36 study cities and contains approximately 5.25 million community–year listing records, including community name, address, monthly average listing price, housing-layout distribution, and construction year. For the same community and month, listing prices from different platforms were cross-validated, and anomalous records with relative deviations greater than 20% were first removed.

Because the community addresses in the publicly available platform records were mostly textual descriptions, such as "No. X, Xierqi North Road, Haidian District, Beijing," and because some historical records had missing or insufficiently accurate longitude–latitude fields, we batch-geocoded all community addresses using the Amap Geocoding API to convert textual addresses into WGS-84 coordinates. For each address, the Amap Geocoding API returns multiple pieces of matching metadata, including match level, administrative-boundary completeness, and match mode. Based on this information, we defined a composite confidence score ($\eta_i$)to quantify the reliability of the geocoding result for the $i$-th community address:

$$\eta_i = w_L \cdot s_L(level_i) + w_A \cdot s_A(admin_i) + w_M \cdot s_M(match_i) \qquad (C4)$$

Here, $s_L(level_i)$denotes the match-level score. Address-level matches were assigned a score of 1.0, POI-level matches 0.95, road-intersection matches 0.85, road-level matches 0.65, township or subdistrict-level matches 0.4, and district/county-level or coarser matches 0.1. $s_A(admin_i)$denotes the administrative-completeness score. A score of 1.0 was assigned when the returned four-level administrative hierarchy—province, city, district, and subdistrict—was fully consistent with the original address description; 0.25 was deducted for each missing or inconsistent administrative level. $s_M(match_i)$denotes the match-mode score, with exact matches assigned 1.0, complete prefix matches 0.8, and fuzzy keyword matches 0.5. The three weights were set to $w_L = 0.5$, $w_A = 0.3$, and $w_M = 0.2$, respectively, reflecting the relative importance of match level, administrative consistency, and match mode in the overall confidence assessment.

After scoring, records with low confidence ($\eta_i < 0.8$), which accounted for approximately 8.4% of the raw records, were manually reviewed. During this review, researchers performed a three-way comparison among the original textual community address, the standardized address and coordinates returned by Amap, and coordinate information for communities with the same name from other platforms. Records with evident ambiguity were flagged or removed. Subsequently, spatial-consistency validation was conducted for all coordinates that passed the initial screening. The coordinates were converted back into administrative text using the Amap reverse geocoding API. If the returned province, city, or district was inconsistent with that reported in the original record, the record was further inspected. Records located outside the administrative boundary of the study area or clearly deviating from the historical locations of the same community—that is, samples with offsets greater than 500 m—were corrected or removed. When coordinate correction was required, the mean coordinates of the same community in adjacent years were preferentially used.

For price quality control, we first constructed annual city-level price anchors using authoritative housing-price indices to cross-calibrate platform-level listing prices. Specifically, multiple sources were integrated, including the National Bureau of Statistics 70-city residential price index, residential sample prices from the China Index Academy, and transaction prices from CRIC. These sources were combined with weights of 0.5, 0.3, and 0.2, respectively, to obtain the annual city-level average price benchmark ($A_{c,y}$). The listing price of each residential community was then compared with the corresponding anchor price for its city and year. Records deviating from the annual city mean by more than 3σ, as well as records for which the cross-platform relative price deviation exceeded 20% for the same community, were treated as outliers and removed. This procedure reduced the influence of inflated listing prices, platform-specific sampling bias, and erroneous data entry.

The raw data contained cases in which records were missing for certain years but were available in both preceding and following years. For example, a community might have valid records in 2017 and 2019 but be missing in 2018. Such cases are more likely to reflect omissions in platform collection than the actual disappearance of the community from the housing market. Therefore, for communities with short-term missing years within their observed life cycle, we applied an imputation strategy based on "city price anchors plus community fixed effects." First, for each community, we estimated its average deviation from the annual city-level mean price across observed years, thereby characterizing its stable premium or discount. The missing-year price was then back-estimated using the city price anchor for the corresponding year. To avoid excessive smoothing in the imputed values, a random perturbation drawn from $N(0, 0.02^2)$in log space was additionally introduced to simulate year-specific independent fluctuations. Communities with more than five consecutive missing years or with observations in only a single year were not imputed and were directly removed. This procedure yielded a community-level housing-price dataset covering 2017–2024 with continuous annual labels. An is_imputed field was retained to distinguish observed values from model-imputed values.

After identity matching, coordinate correction, cross-source validation, outlier removal, and missing-year imputation, we obtained a community-level housing-price label dataset covering 36 study cities over the eight-year period from 2017 to 2024. It should be noted that missing-value imputation was applied only to intermittent missing years within a community's observed life cycle—that is, cases in which valid observations existed in both preceding and following years but were missing in one or a few intermediate years. No extrapolative imputation was performed for communities appearing in only a single year or missing for extended periods, thereby avoiding excessive sample generation. An overview of the cleaned housing-price dataset is provided in Table

C1.

**Table C1. Overview of the processed community-level housing price label dataset.**

| Dimension | Indicator | Value |
|---|---|---|
| Sample size | Number of unique residential communities | 1,802,075 |
| | Final community–year records | 5,482,360 |
| | Average number of covered years per community | 3.04 |
| | Proportion of raw valid observations | 89.90% |
| | Proportion of model-imputed records | 10.10% |
| Temporal coverage | Study period | 2017–2024 (8 years) |
| | Number of communities in 2017 | 495,073 |
| | Number of communities in 2018 | 516,125 |
| | Number of communities in 2019 | 412,684 |
| | Number of communities in 2020 | 681,946 |
| | Number of communities in 2021 | 748,209 |
| | Number of communities in 2022 | 879,316 |
| | Number of communities in 2023 | 884,957 |
| | Number of communities in 2024 | 864,050 |
| Spatial distribution | Number of communities in T1 cities (SHA/PEK/SZX/CAN/CKG/TSN) | 246,846/178,403/118,286/115,861/95,100/60,182 |
| | Median number of communities in T1.5 cities | 48,645 (Nanjing) |
| | Median number of communities in T2 cities | 25,347 (Dalian) |
| | Median number of communities in T3 cities | 19,072 (Urumqi) |
| | Number of communities in the T4 city | 3,231 (Lhasa) |
| Price distribution | Median price across all samples (CNY/m²) | 22,675 |
| | Price range (CNY/m²) | 2,261—267,679 |
| | Mean price range in first-tier cities (CNY/m²) | 33,800—70,500 |
| | Mean price range in third-tier cities (CNY/m²) | 7,800—12,500 |
| | Proportion of extremely low prices (<3,000 CNY/m²) | 0.12% |
| | Proportion of extremely high prices (>200,000 CNY/m²) | 0.01% |

It should be noted that the annual number of residential communities was not artificially constrained to increase monotonically. This is because platform-based listing data are fundamentally market-behavior data rather than administrative census data. Whether a community appears in a given year depends on the availability of valid listing records for that year. Missing records for certain communities in specific years may reflect real market conditions, such as extremely low listing volumes, long periods without transactions, non-inclusion by a platform, or temporary delisting of housing units. They may also arise from technical factors, including changes in historical platform coverage, omissions during data collection, or missing web archives. Therefore, this study applied limited imputation only to short-term gaps within a community's observed life cycle and did not forcibly extrapolate long-term missing periods or years that were never observed. Compared with constructing a fully balanced panel dataset, our approach emphasizes preserving the natural sparsity and temporal heterogeneity of real market observations.

After constructing annual community-level housing-price labels, we further converted them into pixel-level housing-price labels. Because the raw data contain only representative point coordinates for each residential community, rather than its actual parcel boundary, it is inappropriate to simply assign each pixel the price of the nearest community. To improve the spatial continuity of pixel-level labels, we adopted a distance-decay weighted interpolation method. Specifically, for each city $c$, year $y$, and pixel $\pi_k$, we first extracted the set of communities with valid housing-price labels in that city and year, and constructed a KDTree spatial index based on their coordinates. We then searched for the set of nearby communities $N(\pi_k, y)$within a radius $R$and constructed weights according to the distance $d(\pi_k, u)$between pixel $\pi_k$and community $u$:

$$\omega_{k,u} = \frac{\exp(-d(\pi_k, u)/\tau)}{\sum_{v \in N(\pi_k, y)} \exp(-d(\pi_k, v)/\tau)} \quad (C5)$$

Here, $R$denotes the maximum influence radius, beyond which the weight of a residential community is set to zero; $\tau$is the distance-decay parameter, which controls the influence of nearby communities on the estimated pixel-level price. The pixel-level housing-price label is then defined as the weighted average of community prices within the neighborhood:

$$p_{\pi_k, y} = \sum_{u \in N(\pi_k, y)} \omega_{k,u} \, p_{u,y} \quad (C6)$$

If no valid residential community point was found within radius $R$for a given pixel, the nearest top-$K$ communities were used for estimation. In this study, we set $R = 1000$m, $\tau = 300$m, and $K = 5$, ensuring that the labels capture local housing-price variation while preventing distant communities from exerting unreasonable influence on pixel-level prices. Before training, a log1p transformation was applied to Vd15, with units of CNY/m².

**Vd16—Cross-city Functional-zone Anchors**

The Vd16 label consists of cross-city functional-zone anchors manually annotated by the research team. It contains 180 semantic anchors in total, corresponding to 36 cities × 5 functional-zone types: CBD, old urban core, railway station, university, and new district. Each anchor includes three fields: longitude–latitude coordinates, city code, and functional-zone type. The anchors were annotated by manually verifying representative functional-zone locations in each city. For example, CBD anchors may include landmarks such as Beijing Financial Street, Shanghai Lujiazui, and Guangzhou Tianhe CBD, with their WGS84 coordinates recorded.

Vd16 was formulated as a similarity-retrieval task. Given an anchor, the objective was to retrieve the top-k nearest anchors of the same functional type but from different cities among the full set of 180 anchors, while excluding anchors from the same city. The evaluation metrics were Recall@8 and MRR. No probe was trained for this task; instead, ranking was performed directly using cosine similarity between embeddings.

**C.3 Label CSV Format and Quality Control**

All label CSV files were exported using a unified format. The field definitions are provided in Table C2.

**Table C2. Field definitions and data acquisition status codes of the label CSV files**

| Field | Type | Description |
|---|---|---|
| sample_id | int | Row index corresponding to features.npy |
| code | str | One of the 36 city codes |
| year | int | 2017—2024 |
| lon | float64 | WGS84 longitude |
| lat | float64 | WGS84 latitude |
| value__ | float | Label value; the number of dimensions is consistent with the dimensionality of the corresponding label |
| value_status | str | OK/OUT_OF_BBOX/NODATA/NO_FILE |

The value_status field records the data retrieval status of each sampling point. OK indicates successful extraction; OUT_OF_BBOX indicates that the sample point falls outside the spatial extent of the label raster; NODATA indicates that the sample point lies within a raster NoData region; and NO_FILE indicates that the label file for the corresponding year was not found. During evaluation, only samples with value_status=OK were used. Missing samples were excluded on a task-by-task basis when computing evaluation metrics, rather than being removed row-wise across all tasks, thereby maximizing the number of usable samples for each label.

After checking the full dataset, the valid-sample coverage rate for each label, defined as the proportion of samples with value_status=OK among the total 14,400,000 samples, was 100% for Vd1–Vd15. Vd16, the cross-city functional-zone anchor task, is not applicable to this statistic because of its different task formulation; it contains a fixed set of 180 anchors. Overall, the label library satisfies the statistical stability requirements for downstream evaluation.

**Appendix D**

This appendix details the network architectures, loss parameters, optimizer configurations, and evaluation hyperparameters for all models and training protocols used in the empirical diagnostics in §4. All experiments were conducted on 2× NVIDIA RTX 4090 GPUs. The implementation was based on PyTorch 2.x, and the evaluation probes were implemented using scikit-learn 1.4.

**D.1 Shared Configuration**

All methodological ablation experiments in §§4.1.2–4.1.5 were compared under the following unified base configuration. Except for the axis being ablated, all other parameters were held fixed. The shared training configuration is provided in Table D1.

**Table D1. Shared training configuration for the methodological ablation experiments in Sections 4.1.2–4.1.5**

| Item | Configuration |
|---|---|
| Sampling scale | 14,400,000 pixel–year samples (36 cities × 8 years × 50,000 pixels) |
| Optimizer | AdamW ($\beta_1 = 0.9, \beta_2 = 0.999, \epsilon = 10^{-8}$) |
| Learning rate | $1 \times 10^{-3}$ for the base setting and $3 \times 10^{-4}$for text-integration experiments |
| Weight decay | $1 \times 10^{-5}$ for the base setting and $1 \times 10^{-4}$for text-integration experiments |
| Learning-rate scheduler | CosineAnnealingLR, with T_max=epochs |
| Training epochs | 20 epochs; empirically, the validation loss typically reached its optimum between epochs 17 and 19 |
| Batch size | 4,096 |
| Mixed precision | torch.cuda.amp, with FP16 forward computation and FP32 master weights |
| Gradient clipping | Not used |
| Dropout | 0.1, applied between all non-projection layers |
| Random seed | 42, used throughout the entire experimental pipeline |

**D.2 Network Structures of Fusion Architectures (C1–C5)**

C2–C5 share the same per-stream projector, referred to as StreamProjector. Each of the six numerical streams is projected into a 64-dimensional subspace through a two-layer MLP: Linear(ds → 64) → LayerNorm → GELU → Dropout(0.1) → Linear(64 → 64). C1 does not use per-stream projection. The encoder head structures and parameter counts of the fusion architectures are provided in Table D2.

**Table D2. Network structures and parameter counts of the C1–C5 fusion architectures**

| Architecture | Encoder head structure | Number of parameters |
|---|---|---|
| C1 Raw concatenation + shared backbone | Linear(119 → 256) → LayerNorm → GELU → Dropout → Linear(256 → 256) | — |
| C2 Per-stream projection | Linear(384 → 384) → LayerNorm → GELU → Dropout → Linear(384 → 256) | 182,016 |
| C3 Transformer | 2-layer Transformer encoder with 4-head self-attention, hidden dimension of 64, and FFN dimension of 256, followed by MeanPool → Linear(64 → 256) | 112,896 |
| C4 Gating | gate: MLP(119 → 256 → 6) + softmax; head: MLP(64 → 64 → 256) | 49,350 |
| C5 MoE | router: Linear(384 → 4); four experts: Linear(384 → 384) → GELU → Dropout → Linear(384 → 256), with top-$k$ = 2 weighted aggregation | 700,228 |

The numerical-stream models (C1–C5) output embeddings with dimensionality $d = 256$, consistent with the default setting used in the methodological search in §4.1.2. The feasibility of compression to 64 or 128 dimensions is examined separately in §4.1.4.

In the text-integration experiments (§4.1.5, TE1–TE5), the output embedding dimensionality is also set to $d = 256$, so that the projected BGE-M3 embeddings can be aligned in the same 256-dimensional space using the InfoNCE objective. TE1–TE4 add a text projection head on top of C2, implemented as a two-layer MLP ($1024 \rightarrow 512 \rightarrow 256$), while keeping the embedding output dimensionality fixed at 256.

### D.3 Hyperparameters of Self-supervised Training Objectives (G1–G6)

The default hyperparameters for each SSL loss family are listed below. All reconstruction-based losses (G1–G3) share the same decoder hidden dimension of 256, whereas the contrastive losses (G4–G5) share the same temperature parameter $\tau = 0.1$. The key hyperparameters of the self-supervised training objectives are provided in Table D3.

**Table D3. Key hyperparameter settings of the G1–G6 self-supervised training objectives**

| Loss | Key hyperparameters |
|---|---|
| G1 Reconstruction | hidden=256; per_dim_norm=False for the base version and True for the enhanced version |
| G2 Masked reconstruction | hidden=256; mask_prob=0.3, applied independently to each stream |
| G3 Implicit decoding | hidden=256; n_queries=8, denoting the number of randomly queried indices per sample |
| G4 Cross-stream contrastive learning | proj_dim=64; head_dim=128; τ=0.1 |
| G5 Temporal contrastive learning | head_dim=128; τ=0.1 |
| G6 BYOL | proj_dim=128; hidden=256; teacher momentum $m = 0.99$ |

The contrastive projection heads for G4 and G5 are both implemented as two-layer MLPs with the structure $d \rightarrow 64 \rightarrow 128$. In G6, the prediction head of the online network has the same structure as the projection head of the target network, namely $d \rightarrow 256 \rightarrow 128$.

### D.4 Text-integration Configurations (TE1–TE5)

The text-integration experiments build on the optimal numerical recipe identified in §§4.1.2 and 4.1.3, namely C2 + G1 + per_dim_norm with $d = 256$, by adding a text projection head and InfoNCE alignment. The alignment scope, alignment weight, and text projection head structure of each text-access configuration are provided in Table D4.

**Table D4. Configuration of the TE1–TE5 text-access experiments**

| Configuration | Alignment scope | Alignment weight | Text projection head |
|---|---|---|---|
| TE1 | Full 256 dimensions of $z$ | 0.5 | 1024→512→256 |
| TE2 | Full 256 dimensions of $z$ | 0.1 | 1024→512→256 |
| TE3 | Last 64 dimensions of $z$ | 0.5 | 1024→512→64 |
| TE4 | Last 64 dimensions of $z$ | 0.1 | 1024→512→64 |
| TE5 Direct fusion | No alignment | — | 1024→64, followed by concatenation with the six stream-projected vectors to form a 448-dimensional representation |

The total loss is defined as $L = L_{\text{G1}} + w \cdot L_{\text{text}}$, where $L_{\text{text}}$is the bidirectional InfoNCE loss with $\tau = 0.1$. During inference, use_text=False, and text-related gradients are not involved in the forward computation.

### D.5 Dimensionality and Normalization Configurations (§4.1.4)

Three configurations were used to compare embedding dimensionality: M_dim_64 ($d = 64$), M_dim_128 ($d = 128$), and M_dim_256 ($d = 256$). All were based on C2 + G1 + per_dim_norm=True with all streams included.

For the normalization-strategy comparison, per_dim_norm=True applies batch-wise variance normalization independently to each output dimension in the G1 reconstruction loss, whereas per_dim_norm=False applies a unified MSE directly across all dimensions. All other hyperparameters were kept identical. The detailed configurations are provided in Table D5.

**Table D5. Experimental configuration of embedding dimensionality and reconstruction-loss normalization strategies**

| Experimental item | Configuration |
|---|---|
| Embedding-dimensionality comparison | M_dim_64 ($d = 64$), M_dim_128 ($d = 128$), and M_dim_256 ($d = 256$) |
| Base model combination | C2 + G1 + per_dim_norm with all streams included |
| Normalization enabled | per_dim_norm=True, where batch-wise variance normalization is applied independently to each output dimension in the G1 reconstruction loss |
| Normalization disabled | per_dim_norm=False, where a unified MSE is applied directly across all dimensions |
| Controlled variables | All hyperparameters were kept identical except for per_dim_norm |

### D.6 Evaluation Probes (§3.3.3)

**Table D6. Probe configurations for downstream evaluation tasks.**

| Task type | Probe | Configuration |
|---|---|---|
| Regression (Vd1–Vd3 and Vd5–Vd15) | RidgeCV | alphas=[0.01, 0.1, 1.0, 10.0, 100.0], with the optimal value automatically selected using built-in leave-one-out cross-validation |
| Classification (Vd4) | LogisticRegression | max_iter=1000, solver='lbfgs', n_jobs=-1, with L2 regularization |
| Similarity retrieval (Vd16) | Cosine similarity with top-$k$ retrieval | No probe was trained |

### D.7 Training Resources and Runtime

**Table D7. Training resources and runtime of each experimental stage.**

| Stage | Runtime per experiment on a single RTX 4090 GPU |
|---|---|
| Naive baselines BL1–BL5 (Ridge / random encoder) | <10 min |
| Fusion architectures C1–C5, including 20 training epochs | Approximately 30–45 min |
| SSL objectives G1–G6 | Approximately 30–45 min |
| Text-integration configurations TE1–TE5 | Approximately 60–90 min |
| Embedding-dimensionality comparison, M_dim_64/128/256 | Approximately 25–40 min |
| Normalization comparison, M_pernorm_on/off | Approximately 30–45 min |
| Main CAR model | Approximately 60 min |

The complete experimental pipeline in §4, comprising 31 experimental configurations together with the main CAR method and its variants, required approximately 20–30 hours of cumulative training time on 2× RTX 4090 GPUs. All checkpoints and training logs were saved in the runs_p2/{exp_id}/ directory.